\documentclass[final,5p,times,twocolumn]{elsarticle}
\usepackage[colorlinks=true,linkcolor=blue,anchorcolor=blue,citecolor=blue,filecolor=blue,menucolor=blue,runcolor=blue,urlcolor=blue,breaklinks]{hyperref}
\usepackage[utf8]{inputenc}
\usepackage{tabularx}
\usepackage{subcaption}
\usepackage{booktabs}
\usepackage{float}
\usepackage{textcomp}
\usepackage{siunitx}
\usepackage{gensymb}
\usepackage{amsmath,amssymb,amsfonts}
\usepackage{bm}
\DeclareRobustCommand\dashed{\tikz[baseline=-0.6ex]\draw[thick,dashed] (0,0)--(0.54,0);} 

\usepackage{tcolorbox}
\newtcolorbox{llmprompt}[1][]{
  colback=gray!10,
  colframe=black!70,
  boxrule=0.5pt,
  arc=1pt,
  left=6pt,
  right=6pt,
  top=6pt,
  bottom=6pt,
  fonttitle=\bfseries,
  title=#1
}

\newcommand{\vect}[1]{\bm{#1}} 
\newcommand{\matr}[1]{\mathbf{#1}} 
\journal{Elsevier}
\begin{document}

\begin{frontmatter}
    \title{Hybrid Modeling, Sim-to-Real Reinforcement Learning, and Large Language Model Driven Control for Digital Twins}
    \author[label1,label2]{Adil Rasheed\corref{mycorrespondingauthor}}
    {\corref{mycorrespondingauthor}\fnref{fn1}}
    \ead{adil.rasheed@ntnu.no}

    \author[label1]{Oscar Ravik}
    \author[label3]{Omer San}
    \cortext[mycorrespondingauthor]{Corresponding author}
    \affiliation[label1]{
    organization={Department of Engineering Cybernetics, NTNU},
            city={Trondheim},
            country={Norway}
            }
    \affiliation[label2]{
    organization={Mathematics and Cybernetics, SINTEF Digital},
            city={Trondheim},
            country={Norway}
            }
    \affiliation[label3]{
    organization={Department of Mechanical, Aerospace and Biomedical Engineering, University of Tennessee},
            city={Knoxville},
            country={USA}
            }
            
    \begin{abstract}
        This work investigates the use of digital twins for dynamical system modeling and control, integrating physics-based, data-driven, and hybrid approaches with both traditional and AI-driven controllers. Using a miniature greenhouse as a test platform, four predictive models Linear, Physics-Based Modeling (PBM), Long Short-Term Memory (LSTM), and Hybrid Analysis and Modeling (HAM) are developed and compared under interpolation and extrapolation scenarios. Three control strategies Model Predictive Control (MPC), Reinforcement Learning (RL), and Large Language Model (LLM) based control are also implemented to assess trade-offs in precision, adaptability, and implementation effort. Results show that in modeling HAM provides the most balanced performance across accuracy, generalization, and computational efficiency, while LSTM achieves high precision at greater resource cost. Among controllers, MPC delivers robust and predictable performance, RL demonstrates strong adaptability, and LLM-based controllers offer flexible human–AI interaction when coupled with predictive tools.
    \end{abstract}
    

\end{frontmatter}

\section{Introduction}\label{sec:introduction}
In the era of Industry 4.0 \cite{Lasi2014i4} and smart automation, digital twin (DT) \cite{Rasheed2020dtv} has emerged as a promising technology for modeling, monitoring, and control of physical assets. A DT is a virtual representation of a physical asset that synchronizes with its counterpart via sensor data, communication technologies, models, and control algorithms \cite{Stadtmann2023dti}. By enabling remote supervision, predictive maintenance, risk-free experimentation, and continuous optimization without disrupting operations, DTs are especially valuable in safety-critical or resource-intensive settings where candidate control strategies must be validated before deployment \cite{Schluse2018edt}. At its core, a DT couples two indispensable ingredients: (i) predictive models that estimate the system’s future state under given inputs and disturbances, and (ii) control algorithms that convert desired objectives into input sequences that drive the physical system toward those objectives. The fidelity, robustness, and usability of a DT, therefore, rest squarely on the choice and integration of modeling and control approaches.

Historically, two modeling paradigms have dominated: physics-based models (PBMs), built from first principles and mechanistic insight, and data-driven models, which learn input-output mappings directly from measurements (ranging from simple linear regressions to complex recurrent neural networks, and transformers). PBMs provide interpretability and (often) good extrapolation when the underlying physics are well captured; data-driven models can capture unmodeled dynamics and measurement biases but often excel only within the training distribution. Hybrid analysis and modeling (HAM) \cite{San2021haa}, which blends mechanistic structure with data-driven components, is now emerging as a practical alternative aimed at combining the extrapolative strengths of PBMs with the flexibility of learned components. Yet, there is still little consensus about when and how each modeling family should be used in a DT pipeline, or how their strengths trade off between interpolation and extrapolation scenarios in a realistic environment.

On the control side, Model Predictive Control (MPC) and Reinforcement Learning (RL) represent two influential yet distinct philosophies. MPC is attractive because it embeds constraints and optimization directly into the control law, yielding predictable performance when an accurate model and a well-posed optimization problem are available. However, MPC requires substantial domain knowledge (to build a model and formulate costs/constraints) and expertise in numerical optimization to ensure real-time feasibility and stability. RL, by contrast, can discover control strategies without an explicit model; this model-free flexibility comes at the cost of large sample complexity, sensitive reward shaping, and significant tuning and safety concerns when training directly on physical assets. A promising hybrid approach is to use DTs for offline RL training, thereby reducing operational risk during the training; however, the success of such \textit{sim-to-real} transfer critically depends on the predictive fidelity of the DT and on whether learned policies generalize beyond the simulated scenarios used during training. Meanwhile, Large Language Models (LLMs) like ChatGPT \cite{OpenAI2024g4t1}, LLaMA \cite{Touvron2023loa}, and Gemini \cite{Team2025gaf1} have surged into prominence for natural language tasks and for providing human interpretable explanations. Their potential role in control as an interpretable planner, a natural language interface for specifying objectives/constraints, or an explainability layer that makes black-box controllers more transparent is tantalizing but underexplored. 

These observations are expressed in the following research questions that we attempt to address through this work. 

\begin{itemize}
    \item How do HAM, PBM, and DDM compare in predictive accuracy, computational cost, and ability to generalize across interpolation and extrapolation scenarios in a realistic experimental setting ?
    \item To what extent can RL controllers trained offline in a DT be transferred to and reliably operate the corresponding physical system under real-world conditions?
    \item Can LLM-based controllers interpret natural language objectives and constraints to produce safe, explainable, and practically useful control actions, especially when augmented with predictive models ? How does their performance compare to MPC and RL ?
\end{itemize}

Importantly, we do not address these questions through theoretical analysis or simulation alone. Instead, we demonstrate and validate the findings using an experimental setup. Concretely, we implement and compare predictive models (DDM, PBM, HAM) and controllers (MPC, RL, LLM) within a common DT pipeline and then demonstrate their behavior on an actual laboratory scale physical setup. 

The remainder of this article is structured as follows. Section~\ref{sec:theory} explains the physical setup and the theoretical foundations of predictive modeling and control strategies. Section~\ref{sec:methodandsetup} describes the  data generation process and the implementation details of the various models and controllers necessary for the reproduction of the results. Section~\ref{sec:resultsanddiscussions} presents the results. Finally, Section~\ref{sec:conclusions} concludes the article by summarizing the major contributions, and proposing directions for future research.

\section{Theory} \label{sec:theory}
This section describes the experimental physical setup used to evaluate our methods and outlines the theoretical foundations of the key components of this study.

\subsection{Experimental setup}
\label{subsec:experimentalsetup}
\begin{figure}
    \centering
    \includegraphics[width=\linewidth]{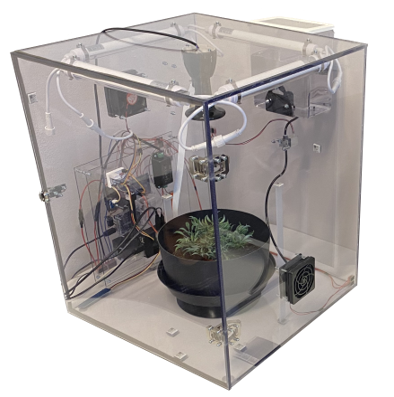}
    \caption{Asset}
    \label{fig:GreenHouseAsset}
\end{figure}

The experimental setup (Figure~\ref{fig:GreenHouseAsset}) used in this study is a miniature greenhouse. The structure is a cuboidal greenhouse measuring $50cm \times 50cm \times 60cm$, constructed using $8mm$ thick UV-resistant polycarbonate sheets. Airtightness is achieved by using solvent welding for the joints and insulation around the hinged door. Environmental control within the greenhouse is facilitated by a set of actuators, including a $H_{max}=100W$ infrared ceramic heater for thermal regulation, full-spectrum LED lights optimized for plant growth, high-flow intake and outtake fans ($F_{max}=68m^{3}/h$) for ventilation and CO$_2$ control, and a water delivery system consisting of a solenoid valve connected to a water tank. 

A sensor suite monitors environmental variables, including temperature, humidity, CO$_2$ concentration, light intensity, soil moisture, and water tank level. Although the focus of the current study is the control of indoor temperature through the regulation of fans and heaters, the presence of a plant contributes meaningful complexity. Its gradual physiological changes introduce natural variability into the environment, serving as a valuable source of uncertainty and enabling us to evaluate the robustness and adaptability of our control models under real-world conditions.  

\subsection{Modeling}
\label{subsec:modelingtheory}
We employ three predictive modeling approaches. First, the PBM derives system dynamics from first principles. Second, the DDM learns system behavior directly from measured data. Third, the HAM integrates well-established physical principles with application-specific data. We now discuss each of these approaches in detail.  

\subsubsection{Physics-Based Modeling}
\label{subsubsec:theorypbm}
Based on the principles of energy conservation, the PBM is given by the following equation:
\begin{equation}
    \frac{dT}{dt} = \frac{H}{\rho V C_p} - \frac{F (T - T_{amb})}{V}
\label{eq:pbm}
\end{equation}
where the term on the left hand side represents the change in inside temperature with respect to time, while on the right hand side, the first term represents the heat input from the heater, where $H$ is the heating power, $\rho$ is the air density, $V$ is the enclosure volume, and $C_p$ is the specific heat capacity of air. Since the heater is operated using a duty cycle $u_h$, $H$ is expressed as $u_{h}H_{max} $. The $u_h$ can be changed in discrete steps of 0.05 from 0-1. The second term represents the heat exchange induced by the fan, where the airflow is modeled as $F = u_{f}F_{max}$, with $u_f$ denoting the fan’s ON/OFF state. $T$ is the internal air temperature, and $T_{amb}$ is the external ambient temperature. 

This model is based on several assumptions. The air inside the enclosure is considered to be perfectly mixed, resulting in a uniform temperature distribution. The properties $\rho$ and $C_p$ are assumed to be constant. The heater and fan are assumed to respond instantaneously, delivering $H$ amount of heat and moving $F$ volume of air per second, respectively. Furthermore, it is assumed that no air flows through the fan when it is off and that no heat is lost through the greenhouse walls. The effects of plants on temperature and flow are also ignored. 
    
\subsubsection{Data-driven modeling}
Two variants of DDM are used: a linear model and a nonlinear LSTM, which are described below.

\paragraph{Linear Model}  
\label{subsubsec:theorylm}  
The system is described by a linear autoregressive model with exogenous inputs (ARX), where the future temperature depends on a finite history of past temperatures and control inputs. The formulation is given by  

\begin{align}
\boldsymbol{x}_{t+1} &= a_1 \boldsymbol{x}_t + a_2 \boldsymbol{x}_{t-1} + \dots + a_p \boldsymbol{x}_{t-p+1} \notag \\
        &\quad + \sum_{j=1}^{q} b_{j}^{(h)} u_{h,t-j+1} 
              + \sum_{j=1}^{q} b_{j}^{(f)} u_{f,t-j+1}
\label{eq:method-arx-equation}
\end{align}  
where $a_i$ are the autoregressive coefficients, and $b_{j}^{(h)}$ and $b_{j}^{(f)}$ represent the influence of the heater duty cycle $u_h$ and fan state $u_f$, respectively. The look-back horizon $p$ captures the temporal dependence of temperature on its past values, while $q$ captures the effect of past control inputs. The state vector and the input vectors are given by $\boldsymbol{x}_t=\left[T_t\right]$ and  $\boldsymbol{u}_t=\left[ u_{h,t}, u_{f,t}\right]^{\top}$. This linear ARX structure enables the prediction of future temperature evolution based on both its recent history and the control input applied to the system.

\paragraph{Long Short-Term Memory}  
\label{subsubsec:theorylstm}  
To capture nonlinear dynamics and long-range dependencies in the system, a data-driven approach based on LSTM \cite{Hochreiter1997lsta} networks is employed. At each time step $t$, the LSTM updates its hidden state $\vect{h_t}$ and cell state $\vect{c_t}$ according to the following equations:  
\begin{align}
    \vect{f_t} &= \sigma \left( \matr{W_f} [\vect{h_{t-1}}, \vect{x_t}] + \vect{b_f} \right) \\
    \vect{i_t} &= \sigma \left( \matr{W_i} [\vect{h_{t-1}}, \vect{x_t}] + \vect{b_i} \right) \\
    \tilde{\vect{c_t}} &= \tanh \left( \matr{W_c} [\vect{h_{t-1}}, \vect{x_t}] + \vect{b_c} \right) \\
    \vect{c_t} &= \vect{f_t} \odot \vect{c_{t-1}} + \vect{i_t} \odot \tilde{\vect{c_t}} \\
    \vect{o_t} &= \sigma \left( \matr{W_o} [\vect{h_{t-1}}, \vect{x_t}] + \vect{b_o} \right) \\
    \vect{h_t} &= \vect{o_t} \odot \tanh(\vect{c_t})
\end{align}  
where $\sigma(\cdot)$ is the sigmoid activation function, $\odot$ denotes element-wise multiplication, $\vect{f_t}$ is the forget gate, $\vect{i_t}$ the input gate, $\vect{o_t}$ the output gate, and $\tilde{\vect{c_t}}$ the candidate cell state.   In this application, the input sequence $\vect{x_t}$ consists of past temperature values and control signals ($u_h$ and $u_f$), while the output is the predicted temperature at the next time step. By leveraging its memory structure, the LSTM can model nonlinear interactions and capture temporal dependencies over longer horizons compared to the linear ARX model.  

\subsubsection{Hybrid Analysis and Modeling}  
\label{subsubsec:theoryham}
The HAM approach used in this work is called the COrrective Source Term Approach (CoSTA), whose theoretical foundation can be found in \cite{Blakseth2022cpb,Blakseth2022dnn,Robinson2022anc}. It augments the PBM introduced in Section \ref{subsubsec:theorypbm} with a data-driven correction term (see Equation~\ref{eq:method-hybrid-model}) to compensate for the modeling assumptions described in Section~\ref{subsec:experimentalsetup}.

\begin{equation}
\frac{d T}{dt} = \frac{H}{\rho V C_p} - \frac{F (T - T_{amb})}{V} + r(T, T_{amb}, H, F; \boldsymbol{\theta}) 
\label{eq:method-hybrid-model}
\end{equation}
Here, the discrepancy between the PBM prediction and the observed data is captured by the corrective source term $r(T, T_{amb}, Q, F; \boldsymbol{\theta})$. Here, $\boldsymbol{\theta}$ denotes the learnable parameters of the DDM used to represent the residual. 

\subsection{Controllers}\label{subsec:controllers}
The study uses three distinct classes of controllers (MPC, RL, and LLM), each of which is summarized in the following sections.

\subsubsection{Model Predictive Control}
\label{subsubsec:theorympc}
MPC is a control strategy that uses a system model to predict the future evolution of the system state, given the current state $\boldsymbol{x}_t$. At each sampling instant, a finite-horizon open-loop optimization problem is solved with the current state as the initial condition $\boldsymbol{x}_0 = \boldsymbol{x}_t$. Only the first control input of the resulting optimal sequence is applied to the system, after which the optimization is repeated at the next time step. In this way, MPC achieves closed-loop behavior by continuously updating the optimization problem with the latest measured state. The finite-horizon optimization problem is formulated as  

\begin{equation}
\begin{aligned}
    \min_{\{\boldsymbol{u}_t\}_{t=0}^{N-1}} \quad & 
    \sum_{t=0}^{N-1} 
    \boldsymbol{x}_{t+1}^\top \mathbf{Q} \, \boldsymbol{x}_{t+1} 
    + \boldsymbol{u}_t^\top \mathbf{R} \, \boldsymbol{u}_t, \\
    \text{s.t.} \quad & \boldsymbol{x}_{t+1} = \mathbf{A}\boldsymbol{x}_t + \mathbf{B}\boldsymbol{u}_t, \\
    & \boldsymbol{u}_{min} \leq \boldsymbol{u}_t \leq \boldsymbol{u}_{max},
    \quad t=0,\ldots,N-1,
    \label{eq:theory-mpc-problem} 
\end{aligned}
\end{equation}

where $N$ is the prediction horizon. The cost function consists of two terms: a state penalty weighted by $\mathbf{Q} \succeq 0$ and a control effort penalty weighted by $\mathbf{R} \succeq 0$. The optimization thus balances driving the state $\boldsymbol{x}$ towards the desired equilibrium (typically the origin) while penalizing excessive control effort. The system dynamics are represented by the linear model of Equation~\ref{eq:method-arx-equation}, and the control inputs are further constrained by lower and upper bounds $\boldsymbol{u}_{min}$ and $\boldsymbol{u}_{max}$.  

\subsubsection{Reinforcement Leanring}
\label{subsubsec:theoryrl}
In RL, an agent interacts with an environment over time in order to learn how to act optimally based on the consequences of its decisions. At each timestep $t$, the agent observes a state $\boldsymbol{x}_t \in \mathcal{S}$, where, in our case, the state corresponds to the temperature $\boldsymbol{x}_t = T_t$. The agent then selects an action $\boldsymbol{a}_t \in \mathcal{A}$; here, the action is composed of the control inputs for heating and flow, $\boldsymbol{a}_t = (u_h, u_f)$.  

After executing an action, the environment evolves to a new state $\boldsymbol{x}_{t+1}$, and the agent receives a reward $r_t$ according to a reward function. In episodic tasks, this interaction continues until a terminal state is reached. The objective of the agent is to maximize the return, defined as the discounted accumulated reward:  
\begin{equation}
    R_t = \sum_{k=0}^{\infty} \gamma^k r_{t+k+1},
\end{equation}
where $\gamma \in (0,1]$ is the discount factor.  

Formally, RL can be described as a Markov Decision Process (MDP) defined by the state space $\mathcal{S}$, action space $\mathcal{A}$, transition dynamics $\mathcal{T}(\boldsymbol{x}_{t+1} \,|\, \boldsymbol{x}_t, \boldsymbol{a}_t)$, and a reward function, together with the discount factor $\gamma$ \cite{Arulkumaran2017drl}. The agent’s behavior is given by a policy $\pi$, which maps states to a probability distribution over actions, $\pi: \mathcal{S} \to p(\boldsymbol{a} \,|\, \mathcal{S})$. The goal is to find the optimal policy $\pi^*$ that maximizes the expected return:  

\begin{equation}
    \pi^* = \arg \max_\pi \, \mathbb{E}[R \,|\, \pi].
\end{equation}
To evaluate policies, a value function $V^\pi(\boldsymbol{x})$ is defined as the expected return starting from state $\boldsymbol{x}$ and following policy $\pi$:  

\begin{equation}
    V^\pi(\boldsymbol{x}) = \mathbb{E}[R \,|\, \boldsymbol{x}, \pi].
\end{equation}
Since the optimal policy $\pi^*$ and optimal value function $V^*$ are generally unknown, a state-action value function, or quality function, is often used:  
\begin{equation}
    Q^\pi(\boldsymbol{x},\boldsymbol{a}) = \mathbb{E}[R \,|\, \boldsymbol{x}, \boldsymbol{a}, \pi],
\end{equation}
which estimates the expected return for taking action $\boldsymbol{a}$ in state $\boldsymbol{x}$ under policy $\pi$. In deep reinforcement learning (DRL), deep neural networks are used to approximate $V^*$ or $Q^*$, enabling the agent to learn optimal control policies directly from data \cite{Mnih2013pawa,Mnih2015hlc}.  In Deep Q-Learning (DQN), a neural network is used to approximate $Q^\pi(\boldsymbol{x},\boldsymbol{a})$ over a discrete action space, allowing the agent to select the optimal control input by choosing the action that maximizes the Q-value:  

\begin{equation}
    \boldsymbol{a}_t = \arg \max_{\boldsymbol{a}} Q^\pi(\boldsymbol{x}_t,\boldsymbol{a}).
\end{equation}

\subsubsection{Large Language Models}
\label{subsubsec:theoryllm}
A foundation model is any model trained on a large dataset that can be adapted to a wide range of downstream tasks \cite{Bommasani2022oto}. One of the most impactful downstream applications has been Natural Language Processing (NLP), where LLMs are now widely deployed. Prominent examples include GPT-4 from OpenAI \cite{OpenAI2024g4t1}, Llama from Meta \cite{Touvron2023loa}, and Gemini from Google \cite{Team2025gaf1}.  LLMs are based on the Transformer architecture \cite{Vaswani2023aia}, which employs attention mechanisms to model dependencies between tokens in the input sequence and the output. They are typically trained in an auto-regressive fashion, generating each token conditioned on the previously generated sequence. This enables them to perform a wide variety of language-related tasks, from text completion to reasoning and planning.  

Beyond pure text generation, LLMs can serve as reasoning and decision-making engines by interacting with external tools. This extends their functionality to domains such as modeling, control, and optimization. For example, an LLM can be prompted with the system state $\boldsymbol{x}_t$ and, depending on the prompt design, decide whether to query a surrogate model for predictions, access a database of historical operating data, or directly issue a control action $\boldsymbol{a}_t = (u_h, u_f)$. In this way, natural language prompts become a high-level interface for connecting system states to external resources and, ultimately, controlling decisions.  

Frameworks such as LangChain \cite{Chase_LangChain_2022} provide orchestration mechanisms that allow LLMs to connect seamlessly to models, databases, or APIs. This enables LLMs to act as controllers that combine symbolic reasoning with data-driven and physics-based information sources. Retrieval-Augmented Generation (RAG) \cite{Gao2024rag} further enhances reliability by allowing the LLM to retrieve relevant knowledge from external sources before generating an output. In control settings, this ensures that actions are informed not only by the pretrained knowledge of the model but also by context-specific data and up-to-date operating conditions.  

By combining tool usage, orchestration frameworks, and retrieval mechanisms, LLMs can operate as hybrid controllers that integrate prior knowledge, predictive models, and real-time reasoning into a unified framework. This approach shifts control from a purely numerical optimization problem to a reasoning-based process in which language interfaces provide a flexible and interpretable means of integrating models, data, and decision logic. Such methods hold particular promise in complex or uncertain environments where adaptability and knowledge integration are essential.

\section{Method and setup} \label{sec:methodandsetup}
This section details the data-generation workflow including preprocessing, splitting strategies, and temperature profiling followed by implementation specifics for modeling and control needed to reproduce the study’s results.

\subsection{Data generation}
The data used for the training and evaluation of the models were generated using the experimental setup described in Section~\ref{subsec:experimentalsetup}, where predefined input sequences of the heating cycle and ON/OFF fan state were applied to the setup. The resulting inside temperature ($T$) measurements were recorded at a sampling interval of $60$ seconds. This procedure yielded $15$ timeseries of internal measurements, each with an average duration of approximately $212$ minutes.

For data-driven time series prediction, the time series was segmented into overlapping input-output pairs using a sliding-window approach. Each input sequence of length $10$ comprised past control inputs and states, while the corresponding output sequence of length $1$ represented the future state to be predicted. The data were partitioned chronologically into training and validation sets using an 80:20 ratio.  

For the HAM approach, the measured dataset was employed to train a deep neural network that estimates the corrective residual term $r(\hat{T}, T_{amb}, H, F; \boldsymbol{\theta})$ in Equation~\ref{eq:method-hybrid-model}. The corresponding features included the current temperature $T$, ambient temperature $T_{amb}$, heater input $u_h$, and fan state $u_f$. The objective of the DNN was to utilize the imperfect PBM prediction of temperature $\hat{T}$, together with $H$ and $F$, to predict the corrective residual term. The dataset was again divided into training and validation subsets using an 80:20 ratio. Further details regarding the implementation of such a HAM model and training strategy are provided in \cite{Blakseth2022dnn}.

Finally, six additional datasets were generated for testing the trained models. The training and testing configurations were designed such that three scenarios correspond to interpolation and three to extrapolation, as summarized in Table~\ref{tab:results-model-scenarios}.
\begin{table}[htbp]
    \caption{Model experiment scenarios.}
    \label{tab:results-model-scenarios}
    \centering
    \begin{tabular}{lll}
        \toprule
        \textbf{Scenario} & \textbf{Traning range} & \textbf{Test range}\\
        \midrule
        1 (Interpolation) & [21.8, 36.9] $\degree C$& [25.5, 30.3]$\degree C$\\
        2 (Interpolation) & [21.8, 36.9] $\degree C$& [25.4, 30.0]$\degree C$\\
        3 (Interpolation) & [21.8, 36.9] $\degree C$& [25.0, 33.3]$\degree C$\\
        4 (Extrapolation) & [21.8, 29.7] $\degree C$& [25.5, 30.3]$\degree C$\\
        5 (Extrapolation) & [21.8, 29.7] $\degree C$& [25.4, 30.0]$\degree C$\\
        6 (Extrapolation) & [21.8, 29.7] $\degree C$& [25.0, 33.3]$\degree C$\\
        \bottomrule
    \end{tabular}
\end{table}
The interpolation Scenarios 1, 2, and 3 represent experiments in which the test temperature profiles fall within the range of the data used to train the models. Specifically, while the models were trained using temperatures in the range of $21.8\degree C-36.9 \degree C$, the test conditions involved temperatures within the range $25.5 \degree C-30.3 \degree C$. This creates a more challenging prediction environment, requiring the models to generalize beyond their learned domain. The extrapolation Scenarios 4, 5, and 6 represent experiments in which the test temperature profiles fall outside the range of the data used to train the models. Specifically, while the models were trained using temperatures up to approximately $29.7\degree C$, the test conditions involved higher target temperatures extending up to $33.3\degree C$. This creates a more challenging prediction environment, requiring the models to generalize beyond their learned domain. For the controller experiments, several temperature profiles were designed as reference trajectories to evaluate the controllers’ ability to regulate the temperature effectively.

\subsection{Predictive models}
\begin{figure}[htbp]
    \centering
    \includegraphics[width=\linewidth]{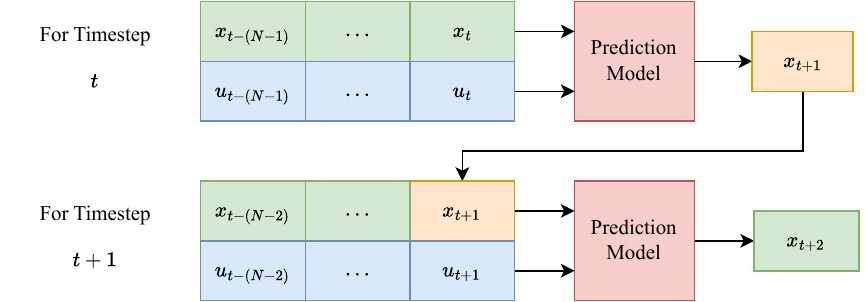}
    \caption[Prediction model experiment architecture]{Architecture for model experiments using the prediction models. For each timestep $t$, the models get the previous $n$ states $\vect{x}$ and the previous and future $n$ controls $\vect{u}$. When the model makes a prediction for timestep $t+1$, it inserts the prediction for timestep $t+1$ into the list of previous measurements before making the next prediction.}
    \label{fig:method-experiments-prediction-model-open-loop}
\end{figure}

For the PBM, Equation~\ref{eq:pbm} was integrated using the RK45 solver from the Python library SciPy. The LSTM and ARX networks predict the system state ($\boldsymbol{x}_{t+1}$) at the next time step ($t+1$), given the previous $N$ states ($\boldsymbol{x}_{t-(N-1)}, \cdots, \boldsymbol{x}_{t}$) and control inputs ($\boldsymbol{u}_{t-(N-1)}, \cdots, \boldsymbol{u}_{t}$). This state ($\boldsymbol{x}_{t+1}$) is then fed back to the model to predict the next state ($\boldsymbol{x}_{t+2}$) and so on to compute the evolution of the system state over a prediction horizon. A schematic is provided in Figure~\ref{fig:method-experiments-prediction-model-open-loop}. The architectural details of the LSTM model and its corresponding hyperparameters are presented in Table~\ref{tab:method-model-parameters-lstm}. To prevent overfitting during training, early stopping was employed. For the HAM model (CoSTA), Equation~\ref{eq:method-hybrid-model} was first solved (similar to the PBM solution explained earlier) without the correction term to obtain the uncorrected temperature $\hat{T}$ at the next time step. The variables $\hat{T}$, $T_{ambient}$, $H$, and $F$ were then used by a feedback neural network to compute the corrective source term $r$. Subsequently, Equation~\ref{eq:method-hybrid-model} was solved again with this corrective term to obtain the correct temperature $T_{t+1}$. The feedforward network architecture and its associated hyperparameters are provided in Table~\ref{tab:method-model-parameters-hybrid}.

\begin{table}
    \caption{Model parameters for the LSTM model.}
    \label{tab:method-model-parameters-lstm}
    \begin{tabular}{l l | l l}
        \toprule
        \textbf{Parameter} & \textbf{Value} & \textbf{Layer} & \textbf{Description}\\
        \midrule
        Epochs & 5000 & LSTM 1         & LSTM\\
        Optimizer & Adam & Linear 1           & Linear\\
        Learning rate & 0.001 & Dropout 1      & Dropout(p=0.2)\\
        Loss function & MSE & LSTM 2         & LSTM\\
        Batch size & 40 & Linear 2           & Linear\\
        Hidden size & 64 & Dropout 2      & Dropout(p=0.2)\\
        Min delta & $5 e^{-4}$ & LSTM 3         & LSTM\\
        Tolerance & 10 & Linear 3           & Linear\\
        \bottomrule
    \end{tabular}
\end{table}

\begin{table}
    \caption{Model parameter for computing the corrective term in the HAM model}
    \label{tab:method-model-parameters-hybrid}
    \begin{tabular}{l l | l l}
        \toprule
        \textbf{Parameter} & \textbf{Value} & \textbf{Layer} & \textbf{Description}\\
        \midrule
        Epochs & 1000 & Linear 1       & Linear\\
        Optimizer & Adam & ReLU 1   & ReLU\\
        Learning rate & 0.001 & Dropout 1      & Dropout(p=0.2)\\
        Loss function & MSE & Linear 2       & Linear\\
        Batch size & 64 & ReLU 2   & ReLU\\
        Hidden size & 64 & Dropout 2      & Dropout(p=0.2)\\
        Min delta & $5 e^{-4}$ & Linear 3       & Linear\\
        Tolerance & 10 & &\\
        \bottomrule
    \end{tabular}
\end{table}

\subsection{Control Algorithms}\label{subsec:control-algorithms}
\begin{figure}[htbp]
    \centering
    \includegraphics[width=1\linewidth]{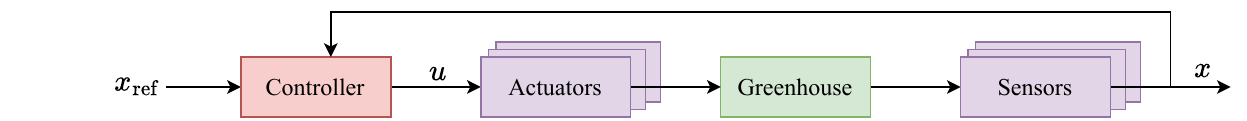}
    \caption[Structure of controller experiment]{Structure for the experiments using the controllers. The controllers have a reference value for the state $\vect{x_{ref}}$, and they get the current state $\vect{x}$ from the sensors. The controllers will send the desired inputs $\vect{u}$ to the actuators, that will affect the temperature in the Greenhouse.}
    \label{fig:method-experiments-controller}
\end{figure}

The overall structure of the full control loop is illustrated in Figure~\ref{fig:method-experiments-controller}. The controller sends a set of control signals to the actuators, which in turn influence the state of the system (greenhouse). The installed sensors then measure the resulting state and transmit this updated information back to the controller. We now provide more details on the three types of controllers in the following sections.

\subsubsection{Model Predictive Control}\label{subsub:mpc}
\begin{figure}[htbp]
\centering
\includegraphics[width=1\linewidth]{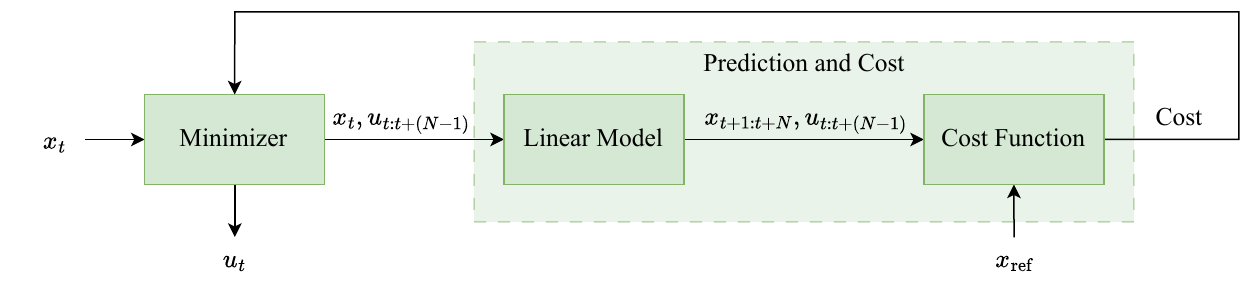}
\caption{Schematic representation of the MPC. The MPC consists of a minimizer that determines the optimal control inputs $\vect{u_t}$ to minimize the cost function. The linear model predicts the current state $\vect{x_t}$ based on the previous state $\vect{x_{t-1}}$ and control inputs $\vect{u_t}$. The cost function evaluates the performance by computing the deviation between the predicted state and the reference state $\vect{x_{ref}}$, as well as the effort associated with the control actions.}
\label{fig:method-MPC-controller}
\end{figure}

Figure~\ref{fig:method-MPC-controller} illustrates the overall structure of the MPC framework. At each control step, the controller predicts the future behavior of the system over a finite horizon using a model of the system dynamics. Based on these predictions, a minimizer selects the sequence of control inputs that minimizes a cost function. This cost function reflects both the deviation of the predicted state from a desired reference and the magnitude of the control effort. Only the first control input from the optimized sequence is applied to the system, after which the process repeats at the next time step with updated measurements. The MPC implementation is based on the linear model described in Section~\ref{subsubsec:theorylm}. The control problem can thus be formulated as a standard linear optimization problem. The optimization objective is defined as

\begin{equation}
    \min\limits_{\vect{u}} \sum_{t=0}^{N-1} Cost(\boldsymbol{x}_t,\boldsymbol{x}_{ref},\boldsymbol{u}_t), \quad s.t \quad \matr{A} \boldsymbol{x}_t + \matr{B} \boldsymbol{u}_t= \boldsymbol{x}_{t+1}, \quad 0 \le \boldsymbol{u} \le 1
    \label{eq:method-mpc-problem} 
\end{equation}
where
\begin{equation}
    \boldsymbol{x} = 
    \begin{bmatrix}
        T \\
    \end{bmatrix}
    ,
    \vect{u} = 
    \begin{bmatrix}
        u_f \\
        u_h \\
    \end{bmatrix}
    \label{eq:method-mpc-state}
\end{equation}
and 
\begin{equation}
Cost(\boldsymbol{x}_t, \boldsymbol{u}_t) = (\boldsymbol{x}_{ref} - \boldsymbol{x}_t)^\top \matr{Q} (\boldsymbol{x}_{ref} - \boldsymbol{x}_t) + \boldsymbol{u}_{k}^\top \matr{R} \boldsymbol{u}_t
\label{eq:method-mpc-cost-function}
\end{equation}
The cost function given by Equation~\ref{eq:method-mpc-cost-function} penalizes deviations from the reference state and large control actions, weighted by the matrices $\matr{Q}$ and $\matr{R}$ in Equation~\ref{eq:method-mpc-cost_matrixes}:

\begin{equation}
    \matr{Q} = 
    \begin{bmatrix}
        w_{T}
    \end{bmatrix}
    ,
    \matr{R} = 
    \begin{bmatrix}
        w_{u_f} & 0 \\
        0 & w_{u_h} \\
    \end{bmatrix}
    \label{eq:method-mpc-cost_matrixes}
\end{equation}
where $w_T, w_{u_f}, w_{u_h}$=$(10,0,0)$ without penalty and $(10,1,1)$ with penalty. When tracking a reference that changes over time, the controller uses only the current reference value when calculating the cost at each step. Thus, the MPC does not anticipate future reference changes within the prediction horizon. In principle, this could be improved by providing the reference corresponding to the future timestep $t+N$, enabling the controller to anticipate upcoming variations. However, since the changes in target temperature during the experiments were relatively small between consecutive timesteps, this adjustment would have little effect.

The optimization problem was solved using SciPy’s \textit{optimize.minimize} function~\cite{Virtanen2020s1f} with the Broyden-Fletcher-Goldfarb-Shanno algorithm. The solver produces continuous-valued control inputs $\boldsymbol{u}$ for both the heater and the fan; because the fan is driven by a relay and therefore requires a binary command, its optimized output is post-processed and rounded to either 0 or 1, while the heater input is rounded to a multiple of 0.05.

\subsubsection{Reinforcement Learning}\label{subsec:rl-controller}
\begin{figure}[htbp]
    \centering
    \includegraphics[width=1\linewidth]{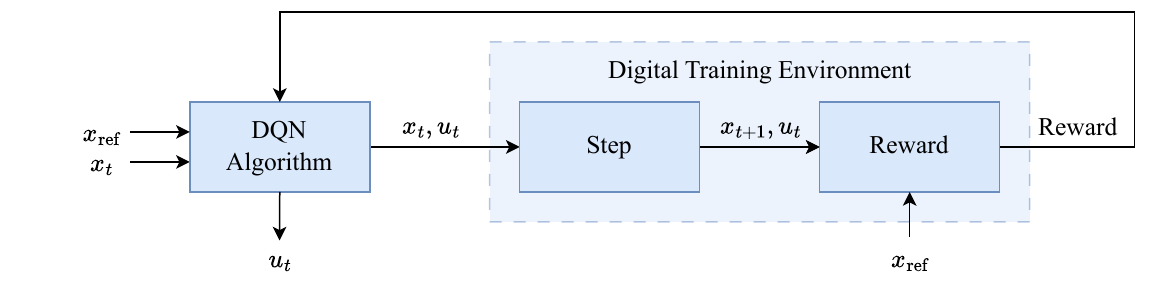}
    \caption[Composition of RL controller]{The composition of the RL controller. It consists of a Deep Q-learning Network algorithm that chooses actions $\vect{u}$ given the current state $\vect{x}$, and the environment that the actions are applied to. The environment will perform a step for each action, that returns the next state. The reward will be caluclated based on the state $\vect{x}$, action $\vect{u}$ and reference state $\vect{x_{ref}}$ in the environment.}
    \label{fig:method-RL-controller}
\end{figure}
\begin{table}
    \caption{RL controller parameters. Abbreviations: Noff – timesteps trained offline, Non – timesteps trained online.}
    \label{tab:model-parameters-rl}
    \centering
    \renewcommand{\arraystretch}{1.2}
    \begin{tabular}{@{}lrrrr@{}}
        \toprule
        \textbf{Parameter} & \textbf{RL-Off-P} & \textbf{RL-Off} & \textbf{RL-On-P} & \textbf{RL-Tr-P} \\
        \midrule
        Noff  & 100\,000 & 100\,000 & 0 & 100\,000 \\
        Non   & 0 & 0 & 1\,440 & 1\,440 \\
        $\lambda_{0f}$ & 0.5 & 0 & 0.5 & 0.5 \\
        $\lambda_{0h}$ & 0.1 & 0 & 0.1 & 0.1 \\
        $\lambda_1$ & 1 & 1 & 1 & 1 \\
        $\lambda_2$ & 0.5 & 0.5 & 0.5 & 0.5 \\
        \bottomrule
    \end{tabular}
\end{table}
The RL controller consists of a Deep Q-learning Network (DQN) algorithm that chooses actions and an environment that models the Greenhouse environment, as shown in Figure \ref{fig:method-RL-controller}. The environment is used to train the DQN algorithm, and it can be trained either online or offline. When the algorithm is trained offline, a model is used to predict the next state during training. During online training, the actions chosen by the algorithm are applied to the actuators in the greenhouse, and the sensors are used to measure the next state. The environment mainly consists of a Step function that performs a step in the environment and a reward function that is used to determine how well the chosen action performed given the current state. The step function receives an action from the algorithm and performs this action on the system. When the agent is trained offline, the environment uses the predictive model (LSTM or HAM) to predict the next state given the current state and future inputs. The reward is computed based on the state of the system, the new state, and the inputs/actions used. The reward function is included in Equation~\ref{eq:method-rl-reward-function}, where the main components are the squared error $||\boldsymbol{x} - \boldsymbol{x}_{ref}||$, the absolute error $|\boldsymbol{x} - \boldsymbol{x}_{ref}|$, and the control inputs $|\boldsymbol{u}|$.  

\begin{equation}
    Reward(\boldsymbol{x}, \boldsymbol{x}_{ref}, \boldsymbol{u}) = -\lambda_2 ||\boldsymbol{x} - \boldsymbol{x}_{ref}|| - \lambda_1 |\boldsymbol{x} - \boldsymbol{x}_{ref}| - \lambda_{0f} |u_f| - \lambda_{0h} |u_h|
    \label{eq:method-rl-reward-function}
\end{equation}
where $\lambda_1, \lambda_2, \lambda_{0f}, \lambda_{0h}$ are weights of the different terms given by Table~\ref{tab:model-parameters-rl}. 

\subsubsection{Large Language Model}\label{subsubsec:method-llm-controller}
\begin{figure}[t]
    \centering
    \begin{subfigure}{0.32\linewidth}
        \includegraphics[width=1\linewidth]{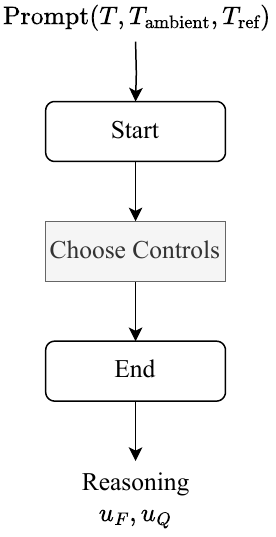}
        \caption{Simple model}
        \label{fig:method-llm-graph-simple}
    \end{subfigure}
    \begin{subfigure}{0.32\linewidth}
        \includegraphics[width=1\linewidth]{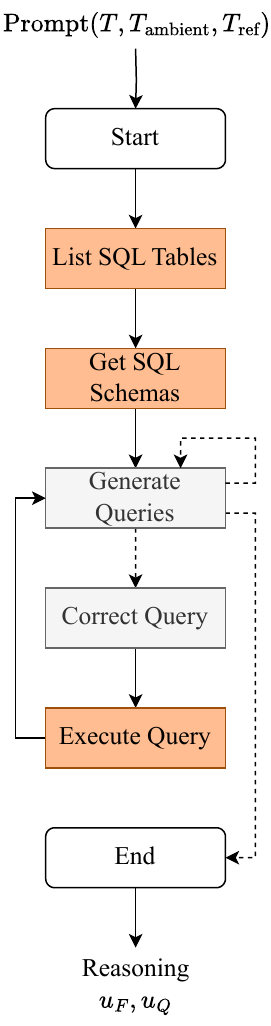}
        \caption{SQL model}
        \label{fig:method-llm-graph-SQL}
    \end{subfigure}
    \begin{subfigure}{0.32\linewidth}
        \includegraphics[width=1\linewidth]{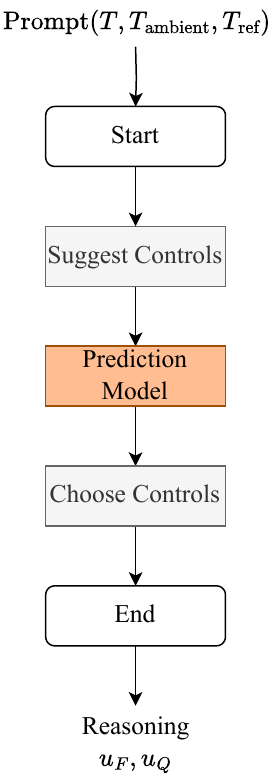}
        \caption{Prediction model}
        \label{fig:method-llm-graph-prediction}
    \end{subfigure}
    \caption[Arcitechture of LLM controllers]{The arkitechture of the three different LLM controller implementations. The rectangles are nodes that are visited in the process, where the gray rectangles consists of an LLM agent, and the orange rectangles are tools the LLM agent can call. The dashed lines are conditional lines, where the action is chosen by the LLM agent based on the current information that the model has access to. Each agent recieves the current temperature $\vect{T}$, the ambient temperature $T_{ambient}$ and the target temperature $T_{ref}$. The outputs from the controller is a reasoning behind the chosen control inputs, and the fan control $u_F$ and heater control $u_H$. }
    \label{fig:method-llm-graph}
\end{figure}

The four variants of LLM-based controllers for regulating temperature evolution were implemented using the Python library \textit{LangChain} and OpenAI’s GPT-4o. \textit{LangChain} enables the construction of tool-augmented agents (e.g., predictive models, SQL databases) capable of executing complex tasks. Each controller outputs $u_f$ and $u_h$, along with an accompanying rationale that explains the selected values. This rationale provides transparency into the controller’s decision-making process.The different implementations are shown in Figure \ref{fig:method-llm-graph}. An example of a prompt that is sent as input to the controller is as follows:

\begin{llmprompt}[LLM Prompt]
    What should the control values heater\_duty\_cycle and fan\_on be set to in order to maintain a temperature of \textit{target temperature} degrees? The temperature now is \textit{current temperature} and the ambient temperature is \textit{ambient temperature} degrees.
    It is important that the temperature in the greenhouse matches the target temperature exactly.
\end{llmprompt}

When the prompt is presented to the model, the placeholders \textit{target temperature}, \textit{current temperature}, and \textit{ambient temperature} are replaced with the corresponding numerical values of the target, current, and ambient temperatures.

The first implementation is a controller that uses the LLM without any tools and obtains the controls through one interaction with the LLM agent. The architecture of the model is shown in Figure~\ref{fig:method-llm-graph-simple}. The second implementation (Figure~\ref{fig:method-llm-graph-SQL}) uses archived historical data to suggest the best possible controls. The third implementation uses predictive models (Linear, HAM and LSTM) to simulate the next timesteps when a set of suggested controls is applied to the prediction model. The LLM models will suggest multiple sets of controls, simulate the next timesteps using the prediction model, and, in the end, use the simulation results to determine the best possible controls. The architecture of the models is included in Figure~\ref{fig:method-llm-graph-prediction}. A penalty for control usage can be introduced by altering the agent's prompt and instructing it to reduce the actuation.  

\subsection{Controller experiments}
The primary objective of the controllers is to maintain a specified reference temperature profile. The controllers are evaluated both with and without control penalties. During the experiment, the controllers determine the control actions required to ensure that the system temperature tracks the reference at each time step. Furthermore, the three controller types (MPC,RL,LLM) are tested under two conditions: with and without penalization of actuator usage. This comparison illustrates the impact of the control penalty on performance. The RL controller is trained offline and deployed for real-time control. This was after realizing that the continued training on the asset provided no improvement in performance. Given the flexibility of the LLM-based controller, several aspects are investigated: the performance of the different implementations shown in Figure~\ref{fig:method-llm-graph}, the effect of control penalties, and the influence of the model’s creativity (temperature) parameter on behavior. 

\subsection{Evaluation criteria}
\label{subsec:evaluationcriteria}
\subsubsection{Model Evaluation criteria}
The relative performance of the models is evaluated based on three main criteria: the Mean Absolute Error (MAE) between the reference and predicted temperatures, the training time, and the memory requirements for loading and running the predictive models. To assess the models, these criteria are applied to both interpolation and extrapolation experiments. Model performance is primarily evaluated using the MAE, while training efficiency and resource usage are compared in terms of total training time and memory footprint, respectively. Memory usage was estimated using the \textit{asizeof} function from the Python library Pympler~\cite{2025p}. Model training was conducted on a MacBook Air equipped with an M2 processor and 16~GB of unified memory. It should be noted that training times may vary depending on the computing hardware used.

\subsubsection{Controller evaluation criteria}
Controller performance is evaluated quantitatively using the Mean Absolute Error (MAE) between the reference and tracked temperature profiles, as well as by the training time required for each controller. We also consider qualitative metrics \textit{implementation effort} and the \textit{level of domain knowledge required} to highlight practical trade-offs among MPC, RL, and LLM controllers.

\section{Results and discussions} \label{sec:resultsanddiscussions}
The following section presents results from the predictive-modeling and control experiments conducted on the greenhouse setup. We evaluate four temperature prediction models (Linear, PBM, LSTM, HAM) under both interpolation and extrapolation scenarios. We then analyze the effectiveness of three control strategies (MPC, RL, LLM) in tracking reference temperature profiles. These experiments provide comparative insights into the accuracy, robustness, and practical feasibility of each modeling and control approach.

    \subsection{Modeling}
    As mentioned earlier, four different kinds of prediction models (Linear, LSTM, PBM, HAM) were developed to estimate the internal greenhouse temperature based on a predefined set of inputs and outputs. To evaluate their performance, six randomly selected scenarios, each corresponding to a distinct target temperature profile, are used for inter model comparison. Among these, Scenarios 1, 2, and 3 represent interpolation cases, while Scenarios 4, 5, and 6 represent extrapolation cases (see Table~\ref{tab:results-model-scenarios}). In all experiments, the models' predicted temperatures are compared to the actual temperatures measured within the physical enclosure under the chosen input conditions. 
    
    \subsubsection{Interpolation}
        \begin{figure}[H]
            \centering
            \begin{subfigure}[b]{\linewidth}
                \centering
                \includegraphics[width=1\linewidth]{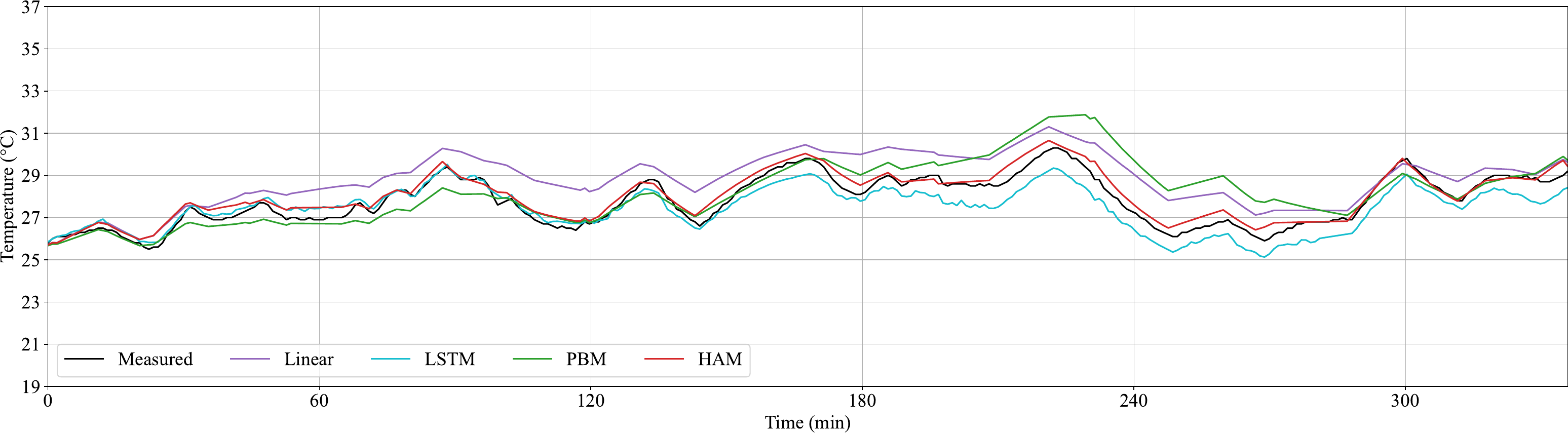}  
                \caption{Temperature}
                \label{subfig:scenario1T}
            \end{subfigure}
            \begin{subfigure}[b]{\linewidth}
                \centering
                \includegraphics[width=1\linewidth]{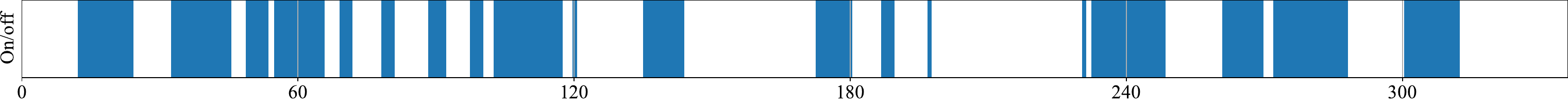} 
                \caption{Fan}
                \label{subfig:scenario1F}
            \end{subfigure}
            \begin{subfigure}[b]{\linewidth}
                \centering
                \includegraphics[width=1\linewidth]{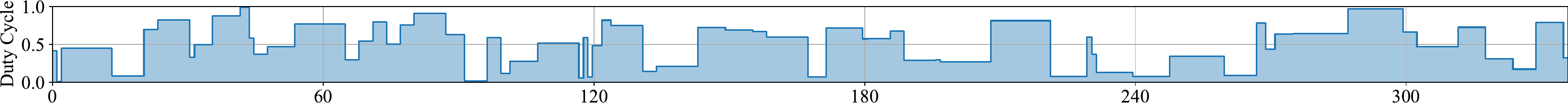} 
                \caption{Heater Duty Cycle}
                \label{subfig:scenario1H}
            \end{subfigure}
            \caption{Model comparison for Scenario 1 corresponding to interpolation.}
            \label{fig:Scenario1}
        \end{figure}
        
        \begin{figure}[H]
            \centering
            \begin{subfigure}[b]{\linewidth}
                \centering
                \includegraphics[width=1\linewidth]{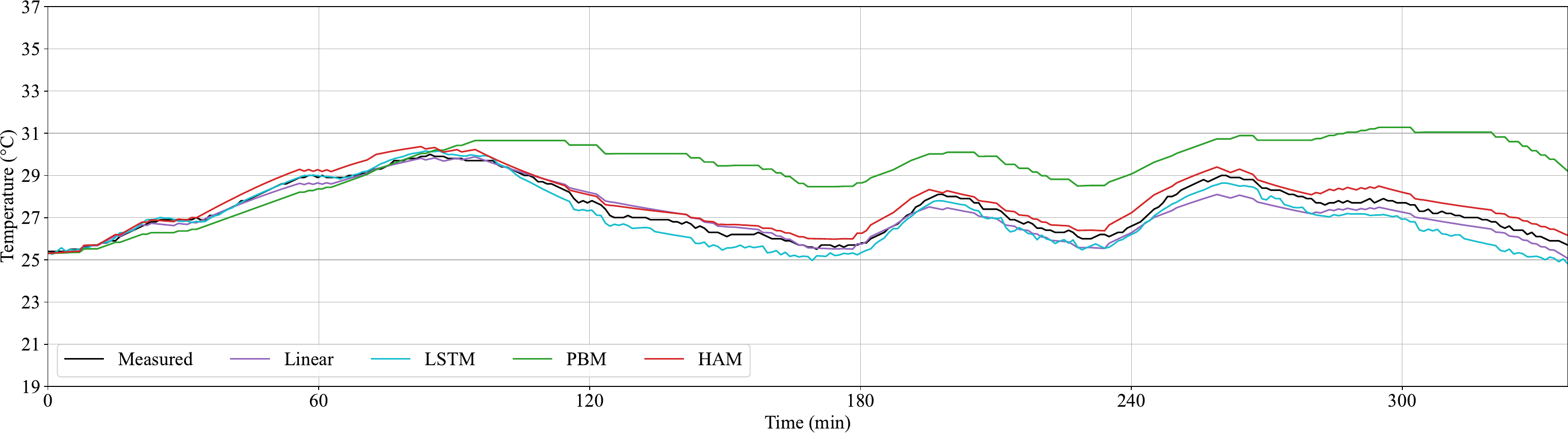}  
                \caption{Temperature}
                \label{subfig:scenario2T}
            \end{subfigure}
            \begin{subfigure}[b]{\linewidth}
                \centering
                \includegraphics[width=1\linewidth]{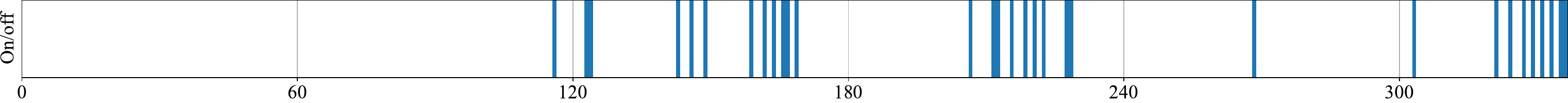} 
                \caption{Fan}
                \label{subfig:scenario2F}
            \end{subfigure}
            \begin{subfigure}[b]{\linewidth}
                \centering
                \includegraphics[width=1\linewidth]{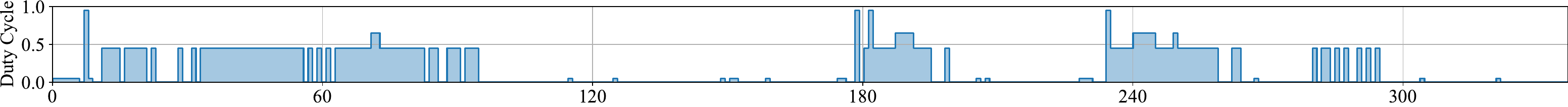} 
                \caption{Heater Duty Cycle}
                \label{subfig:scenario2H}
            \end{subfigure}
            \caption{Model comparison for Scenario 2 corresponding to interpolation.}
            \label{fig:Scenario2}
        \end{figure}
        
        \begin{figure}[h]
            \centering
            \begin{subfigure}[b]{\linewidth}
                \centering
                \includegraphics[width=1\linewidth]{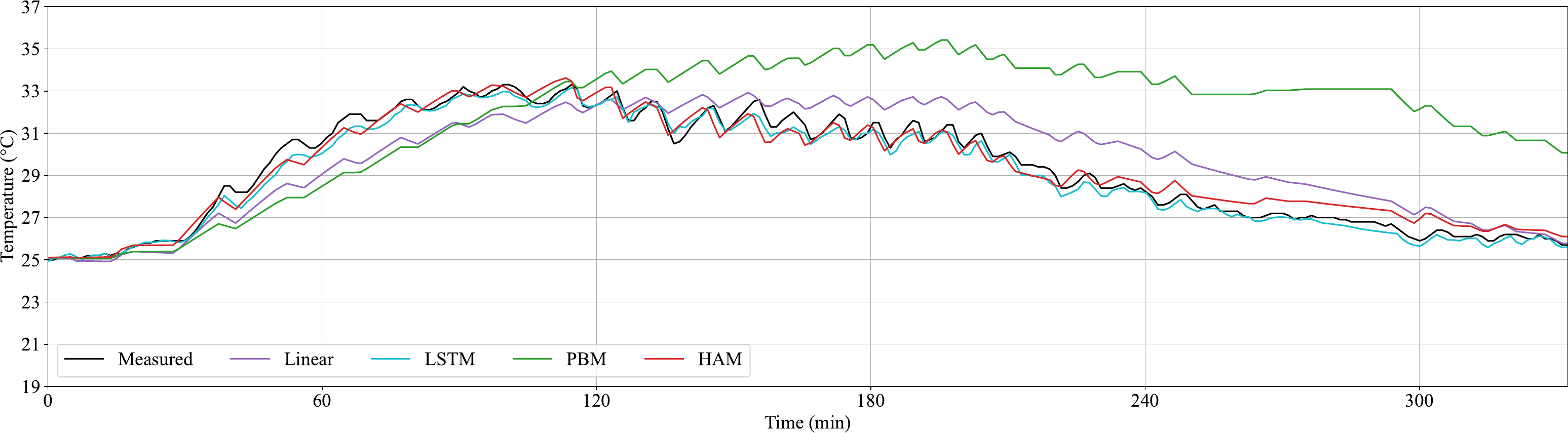}
                \caption{Temperature}
                \label{subfig:scenario3T}
            \end{subfigure}
            \begin{subfigure}[b]{\linewidth}
                \centering
                \includegraphics[width=1\linewidth]{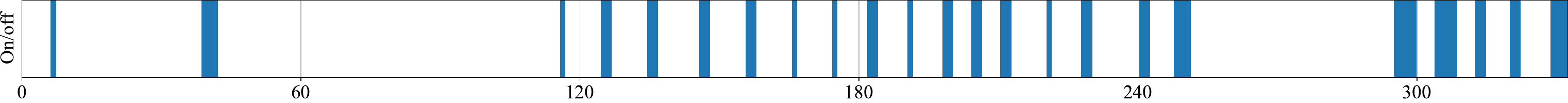} 
                \caption{Fan}
                \label{subfig:scenario3F}
            \end{subfigure}
            \begin{subfigure}[b]{\linewidth}
                \centering
                \includegraphics[width=1\linewidth]{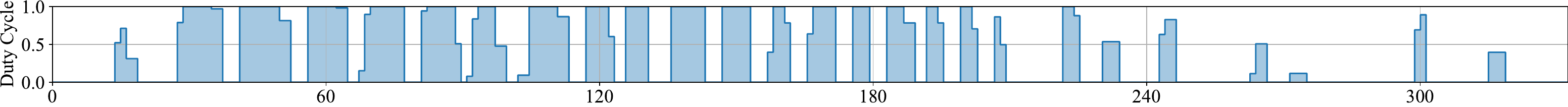} 
                \caption{Heater Duty Cycle}
                \label{subfig:scenario3H}
            \end{subfigure}
            \caption{Model comparison for Scenario 3 corresponding to interpolation.}
            \label{fig:Scenario3}
        \end{figure}
        
        The three interpolation scenarios, 1, 2, and 3, shown in Figures~\ref{fig:Scenario1}, \ref{fig:Scenario2}, and \ref{fig:Scenario3}, respectively, use heater and fan controls to regulate the greenhouse temperature. The main difference between the scenarios lies in how the control inputs, namely the heater duty cycle and fan usage, are varied over time. In Scenario 1 (Figure~\ref{fig:Scenario1}), the heater duty cycle remains non-zero for a larger portion of the time, and the fan is also activated relatively frequently. This consistent control input results in a relatively smooth and well regulated temperature profile with minimal fluctuations. The models, particularly the LSTM and HAM, perform well under these conditions, closely tracking the measured temperature. In Scenario 2 (Figure~\ref{fig:Scenario2}), the heating becomes sparser, but the active heating periods are of longer duration, and the fan is used less frequently. Despite these differences, the temperature profile remains smooth, and the models (except for the PBM) continue to perform accurately. Scenario 3 (Figures~\ref{fig:Scenario3}), however, presents a more challenging control scenario. Both the heater and the fan operate intermittently, leading to more pronounced fluctuations in the output temperature. This makes accurate prediction more difficult, especially for the PBM. These differences highlight how the nature and timing of control inputs can significantly affect both the temperature dynamics and model prediction accuracy, particularly under varying internal heat dissipation and system inertia.
        \begin{figure}[h]
            \centering
            \includegraphics[width=1\linewidth]{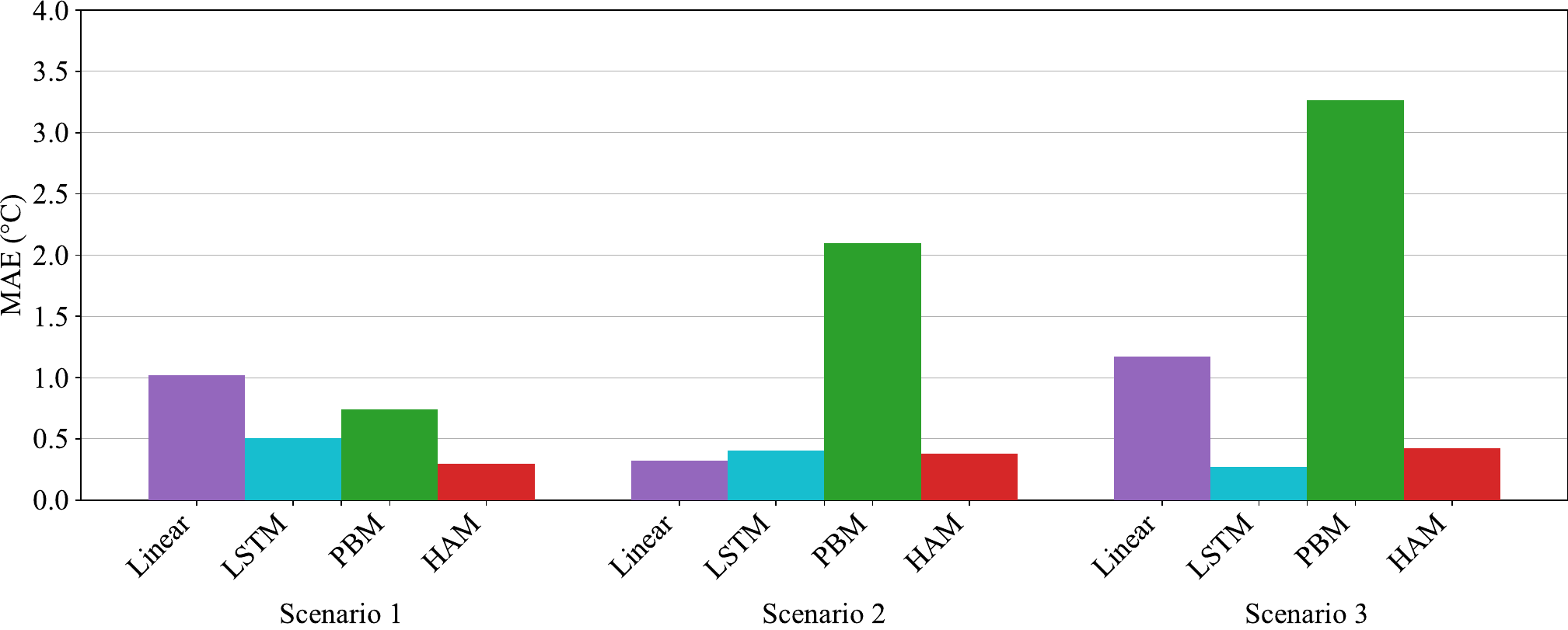}
            \caption{Model comparison for interpolation scenarios.}
            \label{fig:interpolation}
        \end{figure}
        Figure~\ref{fig:interpolation} compares the four different models. The Linear model offers a baseline, performing moderately well under interpolation scenarios. Its simplicity makes it efficient, but it lacks the capacity to model nonlinearities accurately. The PBM, while grounded in physical principles and offering interpretability, falls short in most scenarios. It suffers the consequences of the assumptions mentioned in Section~\ref{subsec:experimentalsetup} leading to relatively poorest performance. The LSTM model, in contrast, demonstrates superior predictive performance across all interpolation scenarios. Due to its ability to model temporal dependencies and non-linearities, it closely follows the measured temperature even under varying heater and fan operations. However, this precision comes at the cost of increased memory usage and longer training times. The HAM strikes a compelling balance. It builds on the physical foundation of the PBM by incorporating a neural network to learn the discrepancies between the model and reality. 
        
        The differences in performance among the models can be largely explained by their structural assumptions and learning mechanisms. The PBM is a first-principles model that does not adapt to the data. Its reliance on simplifying assumptions, such as constant air properties, perfect heat distribution, and negligible unmodeled effects, means that it cannot flexibly capture the complex thermal behavior observed in the greenhouse, particularly under dynamic or extreme conditions. The Linear model, while data-driven, is too simple to handle the nonlinearities of real temperature dynamics; though it does reasonably well in the interpolation scenarios considered here. LSTM and HAM, capable of modeling complex relationships, adapt to measured data effectively. The LSTM benefits from its temporal learning capacity, while the HAM benefits from its hybrid architecture that blends physical understanding with data-driven correction. Together, these characteristics make the LSTM and HAM models more robust, especially in real-world applications where idealized assumptions fail to capture the true behavior of the system.
    
    \subsubsection{Extrapolation}
        \begin{figure}[h]
            \centering
            \begin{subfigure}[b]{\linewidth}
                \centering
                \includegraphics[width=1\linewidth]{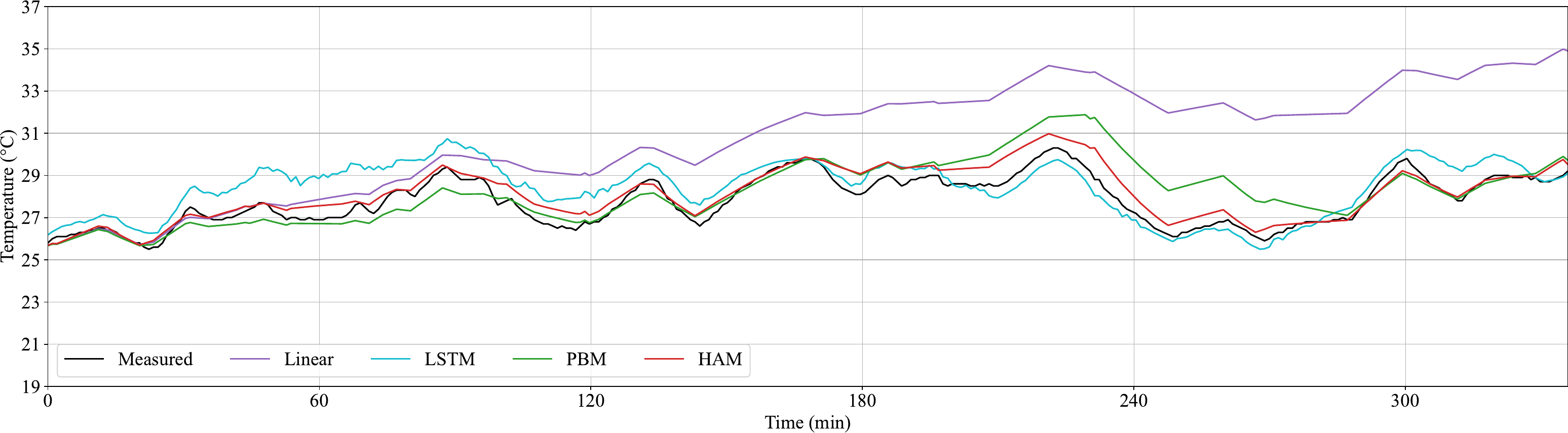}
                \caption{Temperature}
            \end{subfigure}
            \begin{subfigure}[b]{\linewidth}
                \centering
                \includegraphics[width=1\linewidth]{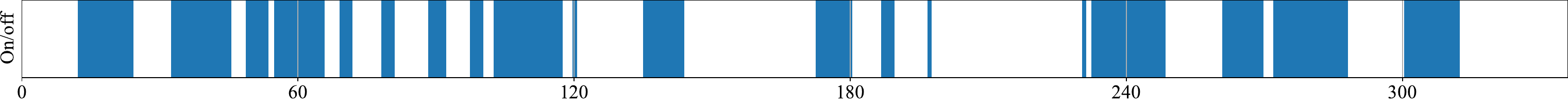}
                \caption{Fan}
            \end{subfigure}
            \begin{subfigure}[b]{\linewidth}
                \centering
                \includegraphics[width=1\linewidth]{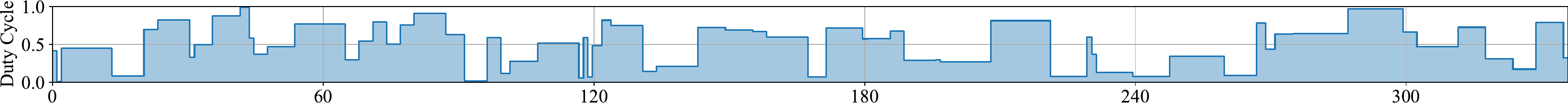}
                \caption{Heater Duty Cycle}
            \end{subfigure}
            \caption{Model comparison for Scenario 4 corresponding to extrapolation}
            \label{fig:Scenario4}
        \end{figure}
        
        \begin{figure}[h]
            \centering
            \begin{subfigure}[b]{\linewidth}
                \centering
                \includegraphics[width=1\linewidth]{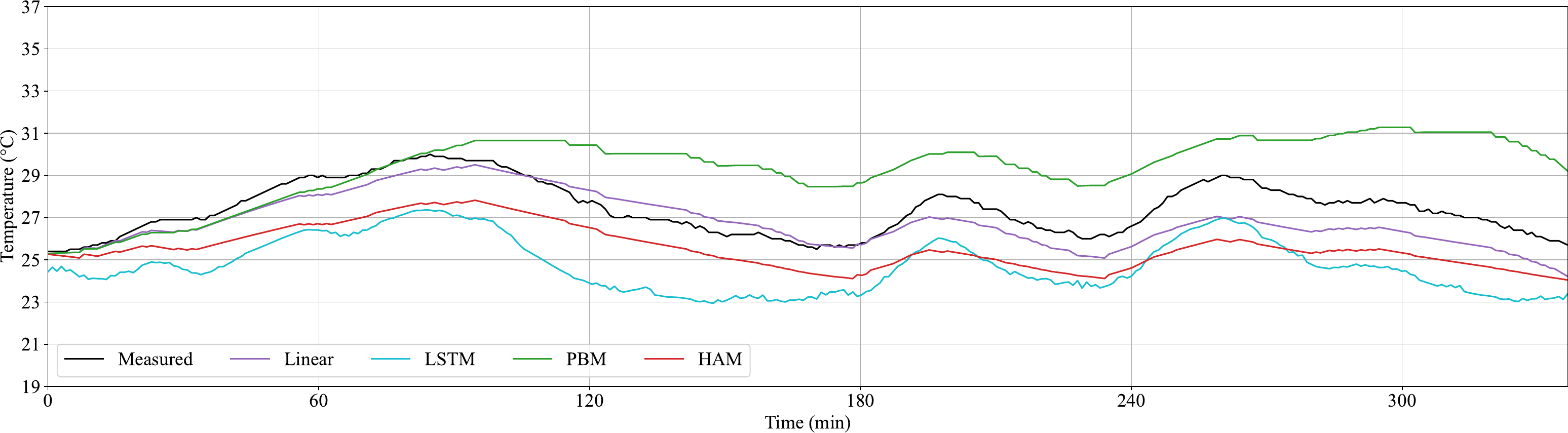}  
                \caption{Temperature}
            \end{subfigure}
            \begin{subfigure}[b]{\linewidth}
                \centering
                \includegraphics[width=1\linewidth]{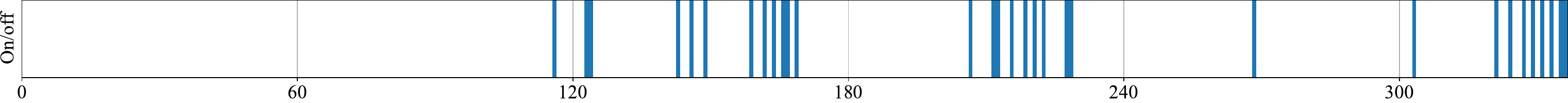} 
                \caption{Fan}
            \end{subfigure}
            \begin{subfigure}[b]{\linewidth}
                \centering
                \includegraphics[width=1\linewidth]{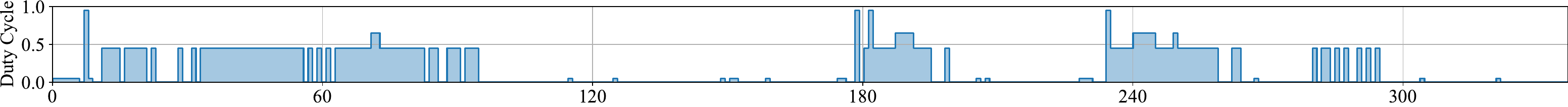} 
                \caption{Heater Duty Cycle}
            \end{subfigure}
            \caption{Model comparison for Scenario 5 corresponding to extrapolation}
            \label{fig:Scenario5}
        \end{figure}
        
        \begin{figure}[h]
            \centering
            \begin{subfigure}[b]{\linewidth}
                \centering
                \includegraphics[width=1\linewidth]{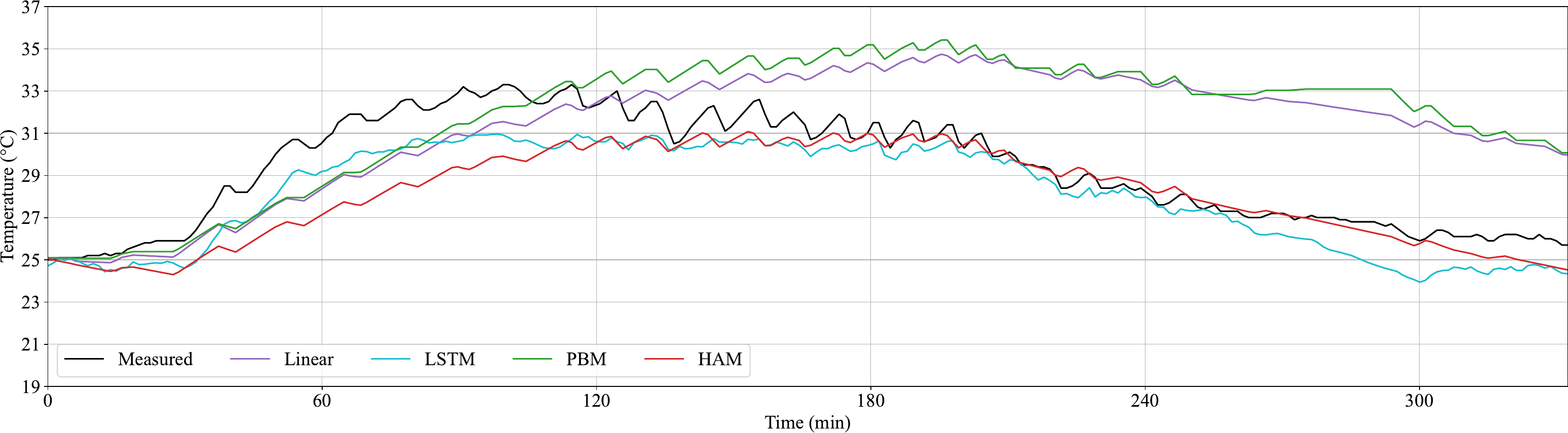}  
                \caption{Temperature}
            \end{subfigure}
            \begin{subfigure}[b]{\linewidth}
                \centering
                \includegraphics[width=1\linewidth]{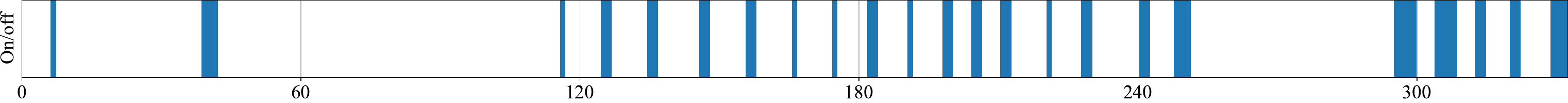} 
                \caption{Fan}
            \end{subfigure}
            \begin{subfigure}[b]{\linewidth}
                \centering
                \includegraphics[width=1\linewidth]{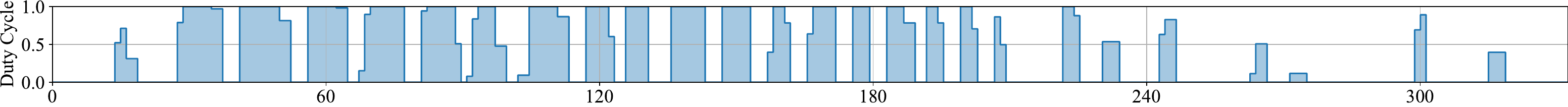} 
                \caption{Heater Duty Cycle}
            \end{subfigure}
            \caption{Model comparison for Scenario 6 corresponding to extrapolation}
            \label{fig:Scenario6}
        \end{figure}
        
        \begin{figure}[htbp]
            \centering
            \includegraphics[width=1\linewidth]{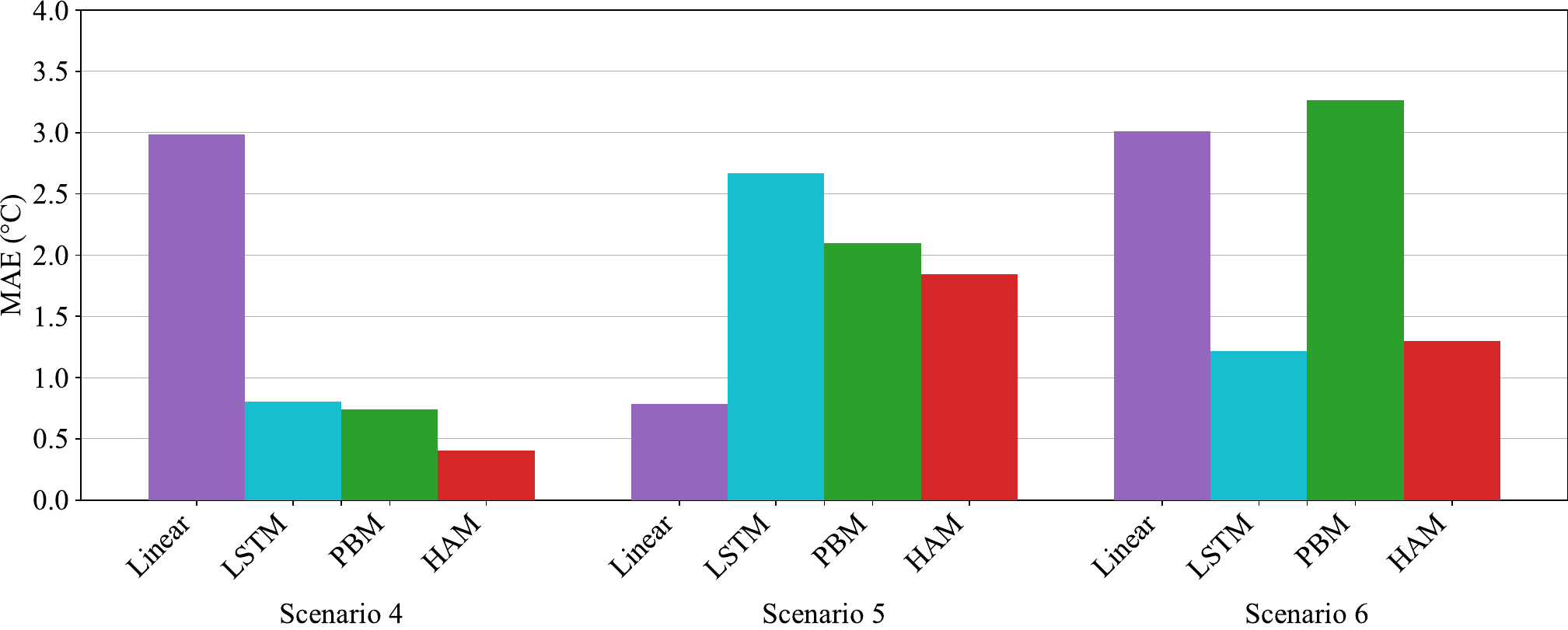}
            \caption{Model comparison for extrapolation scenarios.}
            \label{fig:extrapolation}
        \end{figure}
        The extrapolation scenarios; Scenarios 4, 5, and 6 require the models to generalize beyond their learned scenarios. As expected, the performance of the data-driven models, such as the Linear model, LSTM, and even HAM, deteriorates compared to the interpolation scenarios, as illustrated in Figure~\ref{fig:extrapolation}. The Mean Absolute Error (MAE) increases for all three models across the extrapolation scenarios. However, the extent of this degradation varies with both the model and the specific scenario.
        
        In Scenario 4 (Figure \ref{fig:Scenario4}), which includes moderate temperature variation just outside the training range, the LSTM and HAM models still manage to follow the measured temperature profile reasonably well, although some deviations emerge during periods of rapid temperature change. As shown in the figure, the fan and heater inputs are relatively similar to those seen in the interpolation scenario, but the models struggle to fully capture the delayed thermal response, especially during cooling phases. The PBM, interestingly, maintains a consistent MAE compared to interpolation. This is primarily because it is independent of training data. However, this does not indicate superior adaptability but rather a rigidity that neither improves nor worsens under new conditions. The PBM continues to suffer from the same structural limitations—namely, an inability to capture unmodeled dynamics.
        
        Scenario 5, depicted in Figure \ref{fig:Scenario5}, introduces further stress on the models due to a sparser distribution of control inputs. In this case, the models are required to infer temperature dynamics with less frequent and less diverse actuator signals, particularly heater activity. This results in a sharper increase in error for both LSTM and HAM, as they have not been sufficiently trained on data with such sparse control patterns. The Linear model, too, performs worse here than in Scenario 4. The degradation in performance highlights the limitations of these models when required to extrapolate over both the temperature domain and the distribution of control inputs. Scenario 6, shown in Figure \ref{fig:Scenario6}, reintroduces a wider temperature profile similar to Scenario 3 from the interpolation set, but now in an extrapolation setting. The performance slightly improves compared to Scenario 5, likely because the models encounter a broader distribution of control inputs—providing more cues for inference—even though the temperature still extends beyond the training domain.
        
        The key difference between interpolation and extrapolation becomes clear when analyzing the behavior of the data-driven models. In interpolation, the LSTM and HAM models excel because the test inputs closely match the patterns learned during training. Their predictive strength lies in their ability to recognize and reproduce familiar sequences. However, under extrapolation, their reliance on learned distributions becomes a limitation. They exhibit difficulty adapting to unseen combinations of temperature and control inputs, especially when the control strategy deviates from the training set—for instance, when the heater is used in patterns not previously encountered. This is particularly evident in Scenario 5, where the heater remains inactive for long durations.
        
        Another subtle difference is seen in the consistency of performance. While the LSTM and HAM models show relatively stable behavior across interpolation scenarios, their extrapolation performance is more sensitive to variations in scenario structure. HAM, which augments a physics-based model with a residual learning component, still performs better than the pure LSTM in certain cases. This is likely because the embedded physical laws provide a structural prior that supports generalization. However, even this hybrid approach cannot fully compensate for a lack of exposure to the extrapolated conditions during training.
        
        The performance of the PBM model remains consistently poor across both interpolation and extrapolation. This is expected due to the unmodeled and simplifying assumptions used in the development of the model. The Linear model performed relatively better when the inputs were sparse but suffered when the control input changed constantly.
        
        Overall, extrapolation highlights the limitations of purely data-driven models when applied outside their training boundaries. While HAM offers some mitigation by anchoring predictions in physical laws, it is still fundamentally limited by the scope of their learned corrections.

\subsection{Comparision of MPC, RL and LLM controllers}
\begin{figure}
  \centering
    \includegraphics[width=\linewidth]{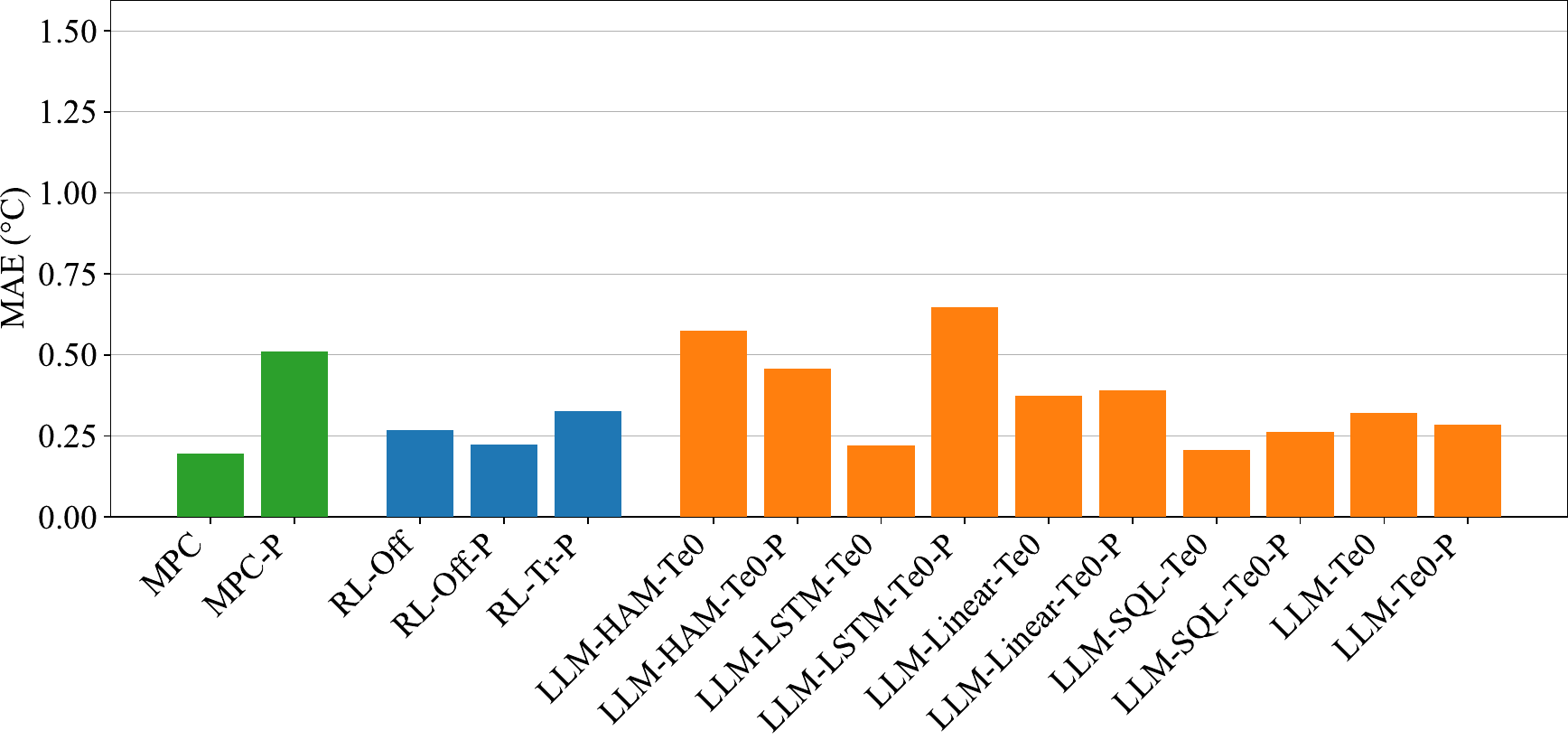}
    \caption{Following reference 2 -P refers to control with penalty}
    \label{fig:results-mae-controllers-2}
\end{figure}

\begin{figure}[t]
    \centering
    \begin{subfigure}[b]{\linewidth}
        \centering
        \includegraphics[width=1\linewidth]{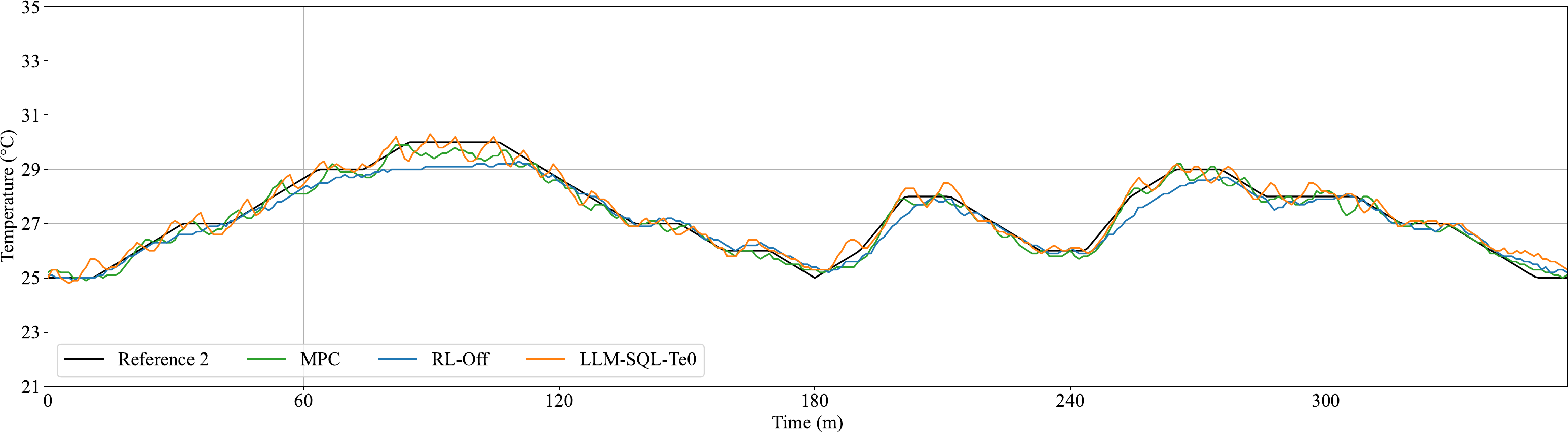}  
        \caption{Temperature}
    \end{subfigure}
    \begin{subfigure}[b]{\linewidth}
        \centering
            \includegraphics[width=1\linewidth]{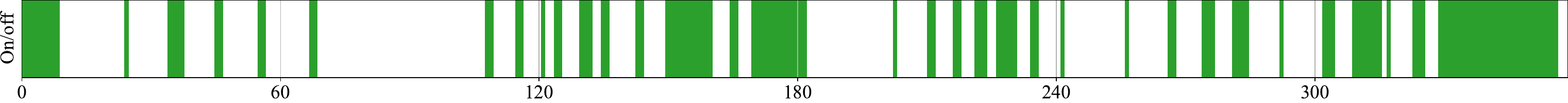}
            \includegraphics[width=1\linewidth]{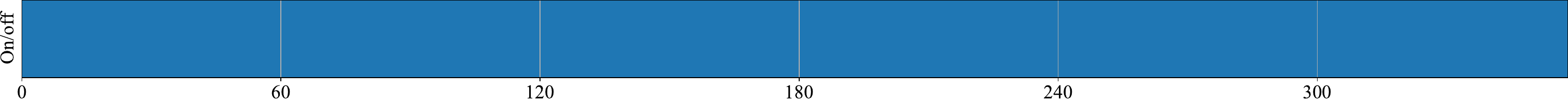}
            \includegraphics[width=1\linewidth]{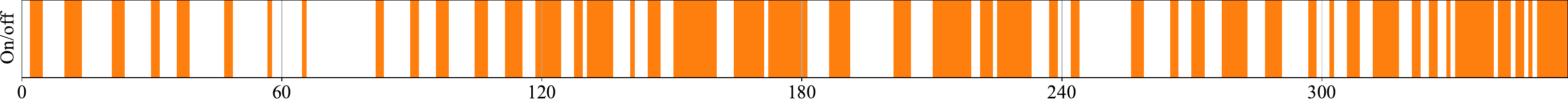}
        \caption{Fan}
    \end{subfigure}
    \begin{subfigure}[b]{\linewidth}
        \centering
        \begin{subfigure}[b]{\linewidth}
            \includegraphics[width=1\linewidth]{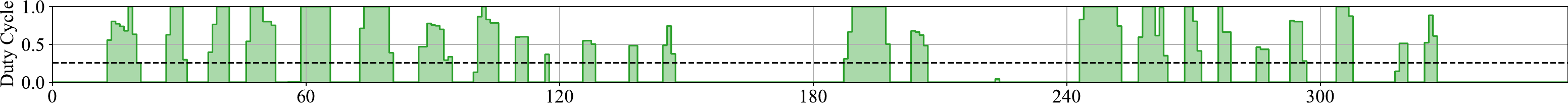}
            \includegraphics[width=1\linewidth]{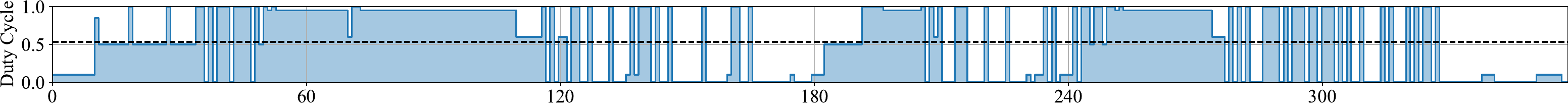}
            \includegraphics[width=1\linewidth]{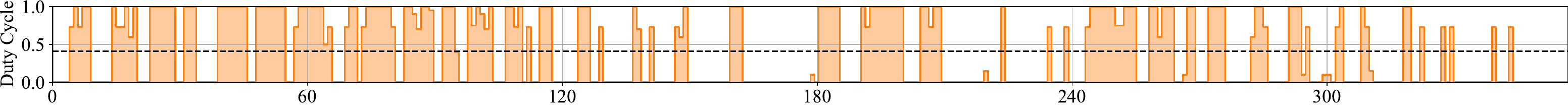}
        \end{subfigure}
        \caption{Heater Duty Cycle, average (\dashed)}
    \end{subfigure}
    \caption{Results from controllers following reference 2 without control penalties. We can see that the RL controller has an active fan during the entire experiment, but is able to follow the reference trajectory quite well. The LLM and the MPC are also able to follow the refeerence quite well, although some oscillations are present.}
    \label{fig:results-controller-no-penalty-2}
\end{figure}

\begin{figure}[t]
    \centering
    \begin{subfigure}[b]{\linewidth}
        \centering
        \includegraphics[width=1\linewidth]{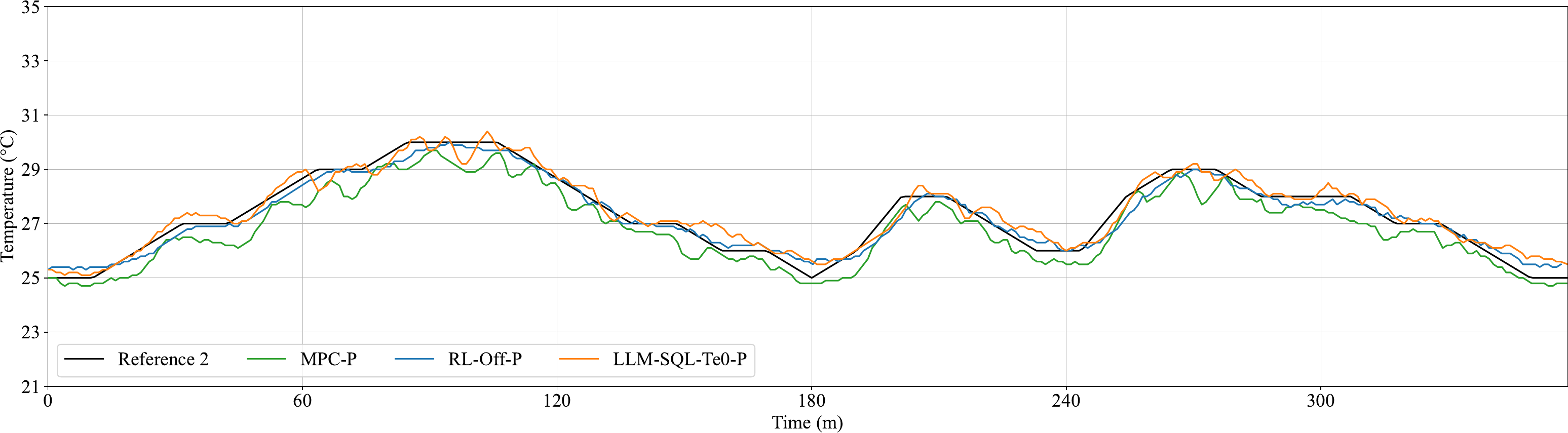}  
        \caption{Temperature}
    \end{subfigure}
    \begin{subfigure}[b]{\linewidth}
        \centering
        \includegraphics[width=1\linewidth]{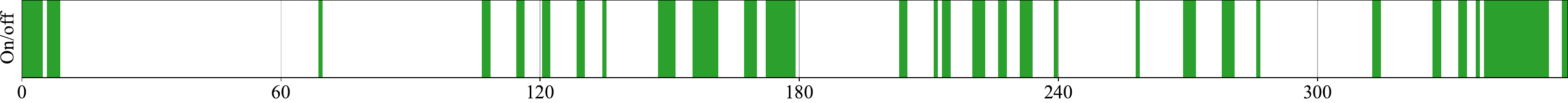}
        \includegraphics[width=1\linewidth]{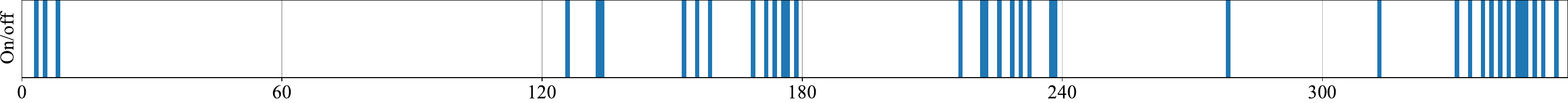}
        \includegraphics[width=1\linewidth]{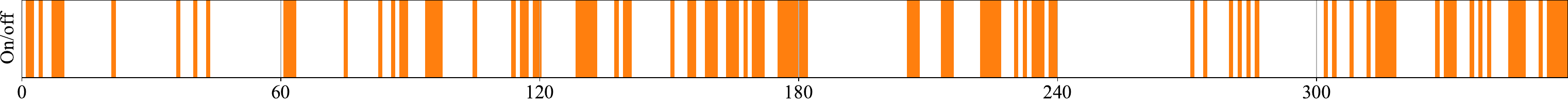}
        \caption{Fan}
    \end{subfigure}
    \begin{subfigure}[b]{\linewidth}
        \centering
        \includegraphics[width=1\linewidth]{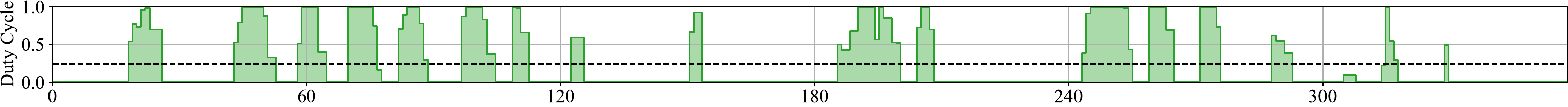}
        \includegraphics[width=1\linewidth]{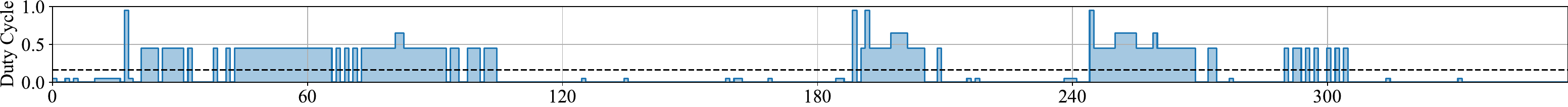}
        \includegraphics[width=1\linewidth]{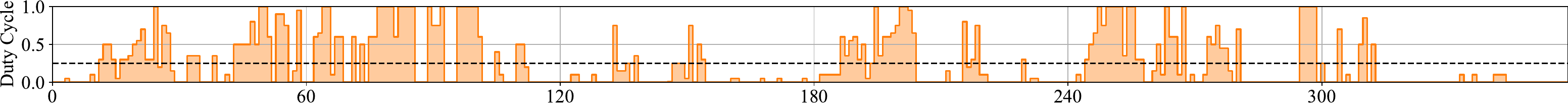}
        \caption{Heater Duty Cycle, average (\dashed)}
    \end{subfigure}
    \caption[Controller with control penalty 2]{Results from controllers following reference 2 with control penalties. We can see that the fan is only active when cooling is required. The MPC struggles with tracking the refereence, as the usage of the heater is penalzed and it is not used enough when needed. The LLM controller is able to follor the reference trajectory quite well. The RL has a smooth trajectory that follows the reference closely. }
    \label{fig:results-controller-penalty-2}
\end{figure}

A comparison of the performance of the MPC, RL, and LLM controllers in tracking a reference temperature profile is summarized in Figures \ref{fig:results-mae-controllers-2}, \ref{fig:results-controller-no-penalty-2}, and \ref{fig:results-controller-penalty-2}.  

Figure~\ref{fig:results-mae-controllers-2} provides a quantitative summary of controller accuracy by presenting the Mean Absolute Error (MAE) for each controller across all experimental runs. Among the three, the MPC demonstrates the lowest MAE. Its design, grounded in optimization over a prediction horizon, allows it to anticipate future states and apply precise corrective actions. The RL controller performs moderately well, with slightly higher MAE values due to its reliance on learned policies that might not have generalized perfectly across all scenarios. The LLM controllers exhibit a wide range of MAE outcomes. Their performance is heavily influenced by the specific architecture and prompting strategies used. Given that the temperature sensor had an uncertainty of $\pm0.5 \degree C$, it is difficult to claim any one model superior to the others. 

Figure~\ref{fig:results-controller-no-penalty-2} illustrates the behavior of the controllers when no penalty is applied to control inputs, meaning that the controllers are free to use the fan and heater without concern for energy efficiency. Under this setting, all controllers focus primarily on minimizing the tracking error. The MPC responds aggressively, frequently toggling the heater and fan to maintain tight adherence to the reference temperature. This results in very close tracking of the reference temperature. The RL controller behaves even more aggressively, guided by a reward function that strongly prioritizes minimizing deviations from the reference temperature. As a result, the fan remains unnecessarily ON for most of the time, which, in turn, requires the heater to operate continuously as well. In the figure, we also present the behavior of the best performing LLM controller. It is clear that it performs very similarly to the MPC controller, albeit utilizing more actuation of the heater and fan. It is interesting to see that the continuous usage of fans resulted in a smoother temperature profile for the RL controller, although it came at the cost of more actuation energy.   

Figure~\ref{fig:results-controller-penalty-2} shows the impact of introducing a control penalty, where actuator usage is explicitly discouraged in the optimization, reward, or prompting process. The effect is evident across all the controllers, as can be seen in a relatively sparser actuation compared to the case without a penalty (Figure~\ref{fig:results-controller-no-penalty-2}). The MPC achieved similar performance even with reduced actuation. The RL controller also adjusted to the penalty, but at the cost of more noticeable degradation in tracking accuracy. The policy learned under penalized conditions struggles to maintain temperature regulation as effectively as it did without the constraint. The LLM controller also exhibited a considerable decrease in actuation without any significant degradation in the tracking error. Interestingly, the MPC controller activates the heater only sparingly; however, but when it does, it typically operates at the maximum duty cycle. In contrast, the RL controller uses the fan exclusively when cooling is required and applies heating at half the permissible duty cycle only when the temperature needs to be raised. In contrast, the LLM controller operated the fan intermittently and adjusted the heater with much greater variation in its duty cycle. 

\begin{figure}[h]
    \centering
    \begin{subfigure}{\linewidth}
        \includegraphics[width=1\linewidth]{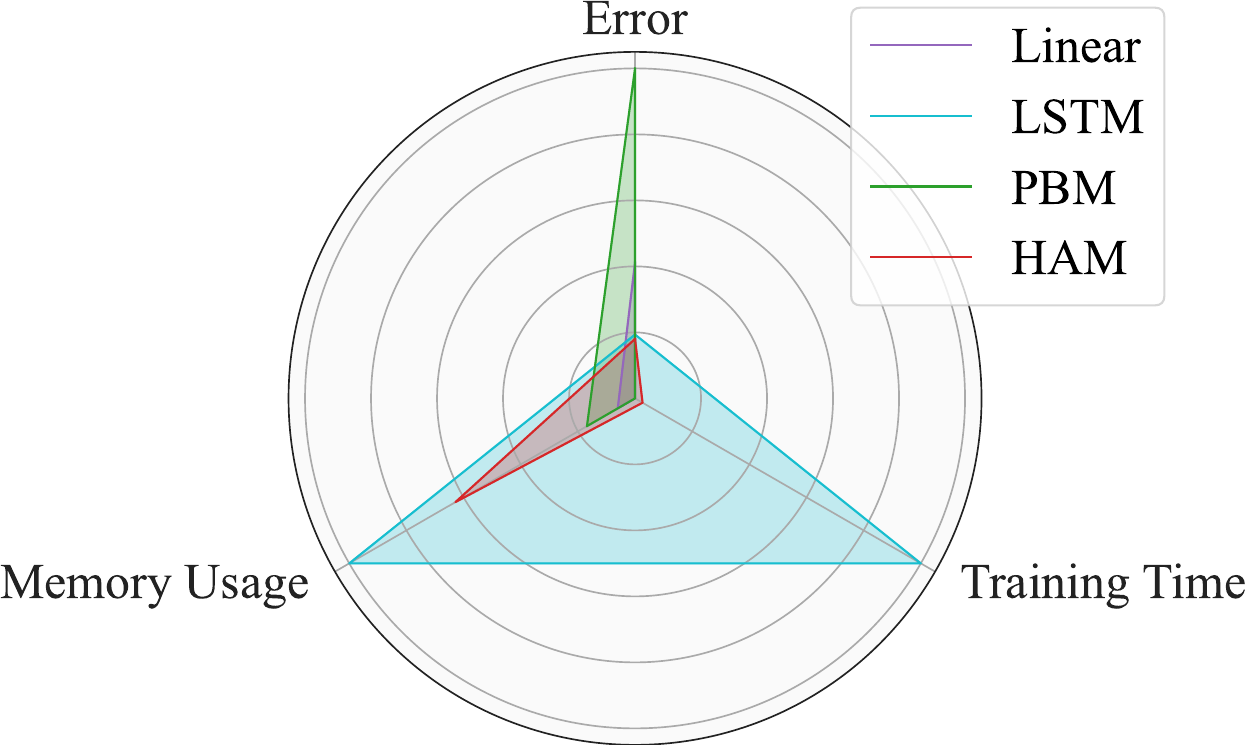}
        \caption{Model}
        \label{subfig:Model-Comparison-Radar-Chart}
    \end{subfigure}
    \begin{subfigure}{\linewidth}
        \includegraphics[width=1\linewidth]{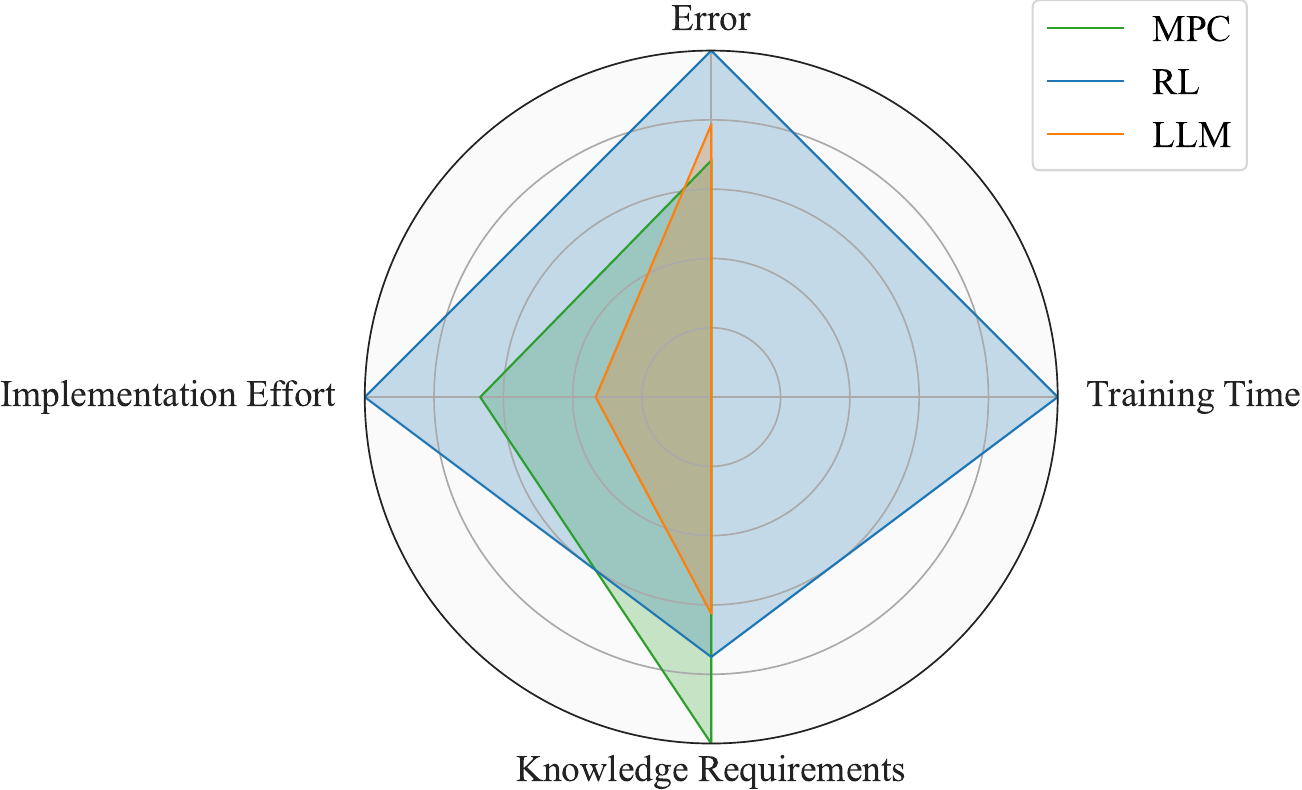}
        \caption{Controllers}
        \label{subfig:Controller-Comparison-Radar-Chart}
    \end{subfigure}
    \caption{An overview of the comparison of models and controllers using quantitative and qualitative matrices.}
    \label{fig:qualquanmodelcontrol}
\end{figure}

\subsection{Reflection on the Models and Controllers Using Quantitative and Qualitative Matrices}
Figure~\ref{fig:qualquanmodelcontrol} presents a comparative analysis of the different models and controllers employed in this work. Figure~\ref{subfig:Model-Comparison-Radar-Chart} compares four modeling approaches based on error, training time, and memory usage. Figure~\ref{subfig:Controller-Comparison-Radar-Chart} compares the three control paradigms in terms of error, training time, implementation effort, and knowledge requirements.

The plot in Figure~\ref{subfig:Model-Comparison-Radar-Chart} highlights clear trade-offs between model accuracy and computational cost.
The LSTM model achieves one of the lowest prediction errors, demonstrating strong temporal learning capabilities and the ability to capture dynamic greenhouse behavior. However, this accuracy comes at the expense of the highest training time and memory usage, indicating significant computational demand. In contrast, the PBM requires no data-driven training, as it relies entirely on first-principles physical relationships. While this leads to minimal computational cost, it results in the largest modeling error, as the simplified physics cannot fully represent real-world nonlinearities. The Linear model performs moderately well, offering faster training and lower memory usage, but its linear assumptions limit its prediction accuracy compared to more expressive models. The HAM provides the best balance among all approaches. By combining physics-based structure with lightweight, data-driven correction, HAM achieves accuracy levels close to those of the LSTM while maintaining far lower computational and memory requirements. This makes HAM an attractive alternative when rapid training or limited computing resources are priorities.

The comparison of controllers in Figure~\ref{subfig:Controller-Comparison-Radar-Chart} underscores differences in training requirements, knowledge dependency, and implementation complexity. RL demands extensive training time and high implementation effort, as it depends on careful reward design, hyperparameter tuning, and large-scale training iterations. Although capable of achieving strong performance, RL requires both computational and methodological expertise. MPC does not rely on data-driven training but requires detailed domain knowledge and expertise in numerical optimization to construct accurate system models and solvers. Once these prerequisites are met, MPC can be implemented efficiently; however, though its performance depends strongly on the quality of the model used. In contrast, the LLM-based controller requires no explicit training phase and minimal domain-specific knowledge. Leveraging reasoning and tool-integration capabilities (e.g., LangChain), the LLM controller can autonomously plan and execute control actions with low implementation effort. Its simplicity and adaptability make it a practical solution for real-time decision making while still achieving competitive control accuracy.

\section{Conclusion and future work} \label{sec:conclusions}
This article explored the use of various predictive models and control strategies to manage the temperature within a greenhouse environment. By integrating physics-based modeling (PBM), data-driven modeling (DDM) approaches, and hybrid analysis and modeling (HAM) approaches, multiple prediction models were developed and tested. These models were then utilized within three distinct control frameworks: Model Predictive Control (MPC), Reinforcement Learning (RL), and Large Language Models (LLMs) to assess their performance in a real-world setup. The study not only evaluated the effectiveness of each controller but also demonstrated how LLMs can be used for intuitive control through natural language. Through this comprehensive investigation, several key findings emerged that highlight both the potential and versatility of modern Artificial Intelligence (AI) techniques in dynamic system control. We itemize these contributions as follows:

\begin{itemize}
    \item \textit{Predictive modeling capabilities:} We found that LSTM and HAM consistently outperformed Linear and PBM across both interpolation and extrapolation regimes. Overall, HAM delivered the best accuracy, maintaining a modest but consistent lead over LSTM and that too with considerably less computational demand.
    \item \textit{Controller capabilities:} We developed and evaluated RL, MPC, and LLM controllers and observed complementary strengths: RL adapted through learning, MPC provided robustness via optimization, and LLMs enabled intuitive and flexible control, particularly when domain expertise was limited.
    \item \textit{Sim-to-real transfer:} We demonstrated effective \emph{sim-to-real} transfer by training RL controllers in the digital twin (DT) and deploying them on the physical setup.
    \item \textit{Natural language for engineering tasks:} We engineered and demonstrated an LLM interface for specifying control logic and constraints.
\end{itemize}

Looking ahead, several promising directions emerge from this work. A key next step is the local deployment of opensource large language models, which would enhance system autonomy, and reduce dependency on external APIs. Future research should also explore enriching LLMs with tool integrations and persistent memory to support more context-aware, sequential decision-making. Additionally, developing a robust automated test framework integrated with version control would ensure consistent validation of experimental changes and support scalability. Exploring other neural architectures and reinforcement learning (RL) algorithms may yield further performance gains, while a focus on natural language interfaces could democratize the development of control systems, empowering users without a technical background to configure and interact with dynamical systems more intuitively. 
\bibliographystyle{AR} 
\bibliography{refs}
\onecolumn
\section*{Appendix}
\begin{table*}[!htbp]
    \caption{Symbols used in controller names.}
    \label{tab:controller-name-notation}
    \centering
    \renewcommand{\arraystretch}{1.2}
    \begin{tabularx}{\linewidth}{p{3.5cm} p{2cm} X}
        \toprule
        \textbf{Symbol} & \textbf{Controller} & \textbf{Penalty}\\
        \midrule
        MPC & MPC & No \\
        MPC-P & MPC & Yes \\
    \end{tabularx}
    \renewcommand{\arraystretch}{1.2}
    \begin{tabularx}{\linewidth}{p{3.5cm} p{2cm} l X}
        \toprule
        \textbf{Symbol} & \textbf{Controller} & \textbf{Penalty} &  \textbf{Training}\\
        \midrule
        RL-Off & RL & No & Offline \\
        RL-Off-P & RL & Yes & Offline\\
        RL-On-P & RL & Yes & Online \\
        RL-Tr-P & RL& Yes & Offline and Online \\
    \end{tabularx}
    \begin{tabularx}{\linewidth}{p{3.5cm} p{2cm} l l X}
        \toprule
        \textbf{Symbol} & \textbf{Controller} & \textbf{Penalty} &  \textbf{Assistance} & \textbf{Temperature}\\
        \midrule
        LLM-Te0 & LLM & No & - & 0\\        
        LLM-Te0-P & LLM & Yes & - & 0\\
        LLM-SQL-Te0 & LLM & No & SQL database & 0\\
        LLM-SQL-Te1 & LLM & No & SQL database & 1\\
        LLM-SQL-Te0-P & LLM & Yes & SQL database & 0\\
        LLM-Linear-Te0 & LLM & No & Linear model & 0\\
        LLM-Linear-Te0-P & LLM & Yes & Linear model & 0\\
        LLM-LSTM-Te0 & LLM & No & LSTM model & 0\\        
        LLM-LSTM-Te0-P & LLM & Yes & LSTM model & 0\\
        LLM-HAM-Te0 & LLM & No & HAM model & 0\\
        LLM-HAM-Te0-P & LLM & Yes & HAM model & 0\\
        \bottomrule
    \end{tabularx}
\end{table*}
\end{document}